\documentclass[10pt,twocolumn,letterpaper]{article}

\usepackage{iccv}
\usepackage{times}
\usepackage{epsfig}
\usepackage{graphicx}
\usepackage{amsmath}
\usepackage{amssymb}
\usepackage{multirow} 
\usepackage{booktabs}
\usepackage{makecell}
\usepackage{enumitem}
\usepackage{titling}
\usepackage[dvipsnames]{xcolor}
\usepackage{algorithm}
\usepackage{algpseudocode}
\usepackage{stmaryrd}
\usepackage{float}

\newcommand{\project}{\textbf{SGAligner}}

\usepackage[pagebackref=true,breaklinks=true,letterpaper=true,colorlinks,bookmarks=false]{hyperref}

\iccvfinalcopy 

\def\iccvPaperID{9305} 

\ificcvfinal\fi

\begin{document}

\title{\textbf{\project{}: 3D Scene Alignment with Scene Graphs}}

\author{
Sayan Deb Sarkar \textsuperscript{1},
Ondrej Miksik \textsuperscript{2},
Marc Pollefeys \textsuperscript{1,2},
Daniel Barath \textsuperscript{1},
Iro Armeni \textsuperscript{1}
\and 
\textsuperscript{1}{\normalsize Department of Computer Science, ETH Zurich, Switzerland} \\
\textsuperscript{2}{\normalsize Microsoft Mixed Reality \& AI Lab, Zurich, Switzerland} \\
{\tt\small \href{https://sayands.github.io/sgaligner/}{\textcolor{blue}{sgaligner.github.io}}}
}
\date{}

\maketitle
\ificcvfinal\thispagestyle{empty}\fi

\begin{abstract}
   Building 3D scene graphs has recently emerged as a topic in scene representation for several embodied AI applications to represent the world in a structured and rich manner. With their increased use in solving downstream tasks (\eg, navigation and room rearrangement), can we leverage and recycle them for creating 3D maps of environments, a pivotal step in agent operation? We focus on the fundamental problem of aligning pairs of 3D scene graphs whose overlap can range from zero to partial and can contain arbitrary changes. 
   We propose \project{}, the \textit{first} method for aligning pairs of 3D scene graphs that is robust to in-the-wild scenarios (\ie, unknown overlap -- if any -- and changes in the environment). We get inspired by multi-modality knowledge graphs and use contrastive learning to learn a joint, multi-modal embedding space. We evaluate on the 3RScan dataset and further showcase that our method can be used for estimating the transformation between pairs of 3D scenes. Since benchmarks for these tasks are missing, we create them on this dataset. The code, benchmark, and trained models are available on the project website.
\end{abstract}

\vspace{-5pt}
\section{Introduction}
\label{sec:intro}
Generating accurate 3D maps of environments is a key focus in computer vision and robotics, being a fundamental component for agents and machines to operate within the scene, make decisions, and perform tasks. 
As such, these maps should be actionable, \ie, containing information (such as objects, instances, their position, and relationship to other elements) that allows agents to perform an action  and represented such that it is easily scalable, updateable, and shareable. 
Recently, 3D scene graphs \cite{armeni20193d,3dssg,rosinol20203d,kim20193} have emerged as a topic in scene representation, providing a structured and rich way to represent the world. 
Not only do they fit the above requirements, but they can also be a lighter-weight~\cite{dlite} and more privacy-aware representation of 3D scenes than the predominantly used 3D point clouds or voxel grids -- hence being easier and safer to share across agents and humans operating in the same scene \cite{li2022remote,zhangdual}. 

Given their potential, 3D scene graphs are increasingly used in embodied agents as a representation -- commonly built on the fly -- to perform robotic navigation~\cite{sepulveda2018deep,rosinol2021kimera,9812179,li2022remote,dlite} and task completion~\cite{Gadre_2022_CVPR,pmlr-v164-agia22a,ravichandran2020bridging,jiao2022sequential,li2022embodied}. Since more and more agents are already building 3D scene graphs for downstream tasks, we investigate how to leverage and recycle them for creating 3D maps of the environment -- a pivotal step in the agent operation -- directly on the scene graph level. Specifically, we examine the fundamental problem of aligning partial 3D scene graphs of an environment that originate from different observations. We focus on real-world scenarios and specifically formulate the problem as follows: given two 3D scene graphs that represent 3D scenes whose overlap can range from zero to partial or full and can contain changes, our goal is to find an alignment across nodes, if it exists. Interestingly enough, even though entity alignment (\ie, node alignment) is used in knowledge graphs and in linguistics \cite{eva,hmea,mmkg,mkhan,mmea,cheng2022multijaf,masked-mmea,mclea}, the task of aligning 3D scene graphs of environments has not been explored. An important note is that entity alignment in these domains assumes that there is overlap between graphs and that all inputs contain true information.

\begin{figure}
    \centering
    \label{fig:teaser}
    \includegraphics[width=1.0\columnwidth]{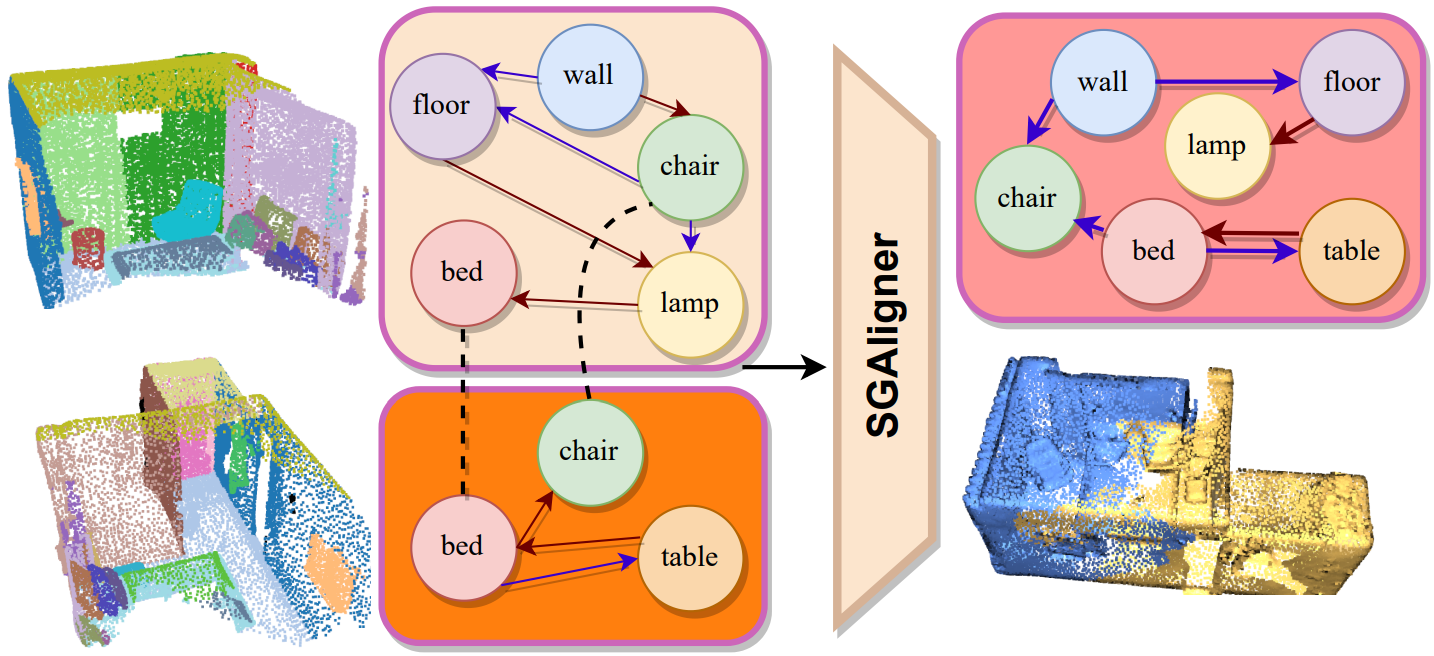}
    \caption{\textbf{\project{}.} We address the problem of aligning 3D scene graphs of environments using multi-modal learning and leverage the output for the downstream task of 3D point cloud registration. Our approach operates directly on 3D scene graph level, is fast and robust to real-world scenarios.} 
    \vspace{-12pt}
\end{figure} 

We propose \project{}, the \textit{first} method for aligning pairs of 3D scene graphs that is robust to in-the-wild scenarios (\ie, unknown overlap -- if any -- and changes in the environment) (see Figure~\ref{fig:teaser}). 
We get inspired by entity alignment methods in multi-modality knowledge graphs \cite{mclea} and redesign them for our setting. 
3D scene graphs represent three main types of information \cite{3dssg,armeni20193d,rosinol20203d}: 
semantic entities in the scene (\eg, object instances), 
their attributes (\eg, category, size, and material), 
and relationships between the entities (\eg, relative position and attribute similarity). 
The main premise is to independently encode each of these modalities with the ultimate objective of learning a joint embedding that can reason how similar two nodes are. 
Given node matches, we perform the scene graph alignment using the matches with the highest similarity. 

We additionally demonstrate our scene graph alignment method on the tasks of 3D point cloud registration and 3D alignment of a local 3D point cloud on a larger map that contains changes.
Instead of directly computing 3D correspondences on the entire point clouds \cite{predator,geotr,regtr,d3feat}, we use the alignment as coarse initialization for the registration. 
We further refine it by computing 3D correspondences \cite{geotr} on the individual point clouds (\ie, object instance point clouds) of each matched node pair. 
This is followed by robustly estimating \cite{martin1981random} the rigid point cloud transformation using the correspondences from all matched nodes. 

We evaluate all three tasks on the 3RScan \cite{3rscan} dataset, which contains 3D point clouds captured over time along with their 3D scene graph annotations \cite{3dssg}. Since 3RScan does not provide partial scene graphs of the same scene or a point cloud registration benchmark and there is no 3D scene graph alignment benchmark, we create the data, metrics, and evaluation needed for these tasks in 3RScan.  
Our experiments show that our approach reduces the relative translation error of state-of-the-art GeoTransformer~\cite{geotr} by $40\%$ in point cloud registration, while being $3\times$ faster during the overlap check, since it does not need to process the entire point clouds. Detailed ablation studies, along with experiments on the task of aligning a changed local 3D scene to a prior 3D map, demonstrate robustness of our approach.
\vspace{5pt}

\noindent We summarize the contributions of this paper as follows: \vspace{-6pt}
\begin{itemize}[leftmargin=8pt,itemsep=-3pt]
    \item We propose \project{}, the first method for aligning pairs of 3D scene graphs whose overlap can range from zero to partial and that may contain changes.
    \item We demonstrate the potential of our method on the tasks of 3D point cloud registration, 3D point cloud mosaicking and 3D alignment of a point cloud in a larger map that contains changes.
    \item We create a scene graph alignment and 3D point cloud registration benchmark on the 3RScan \cite{3rscan,3dssg} dataset, with data, metrics, and evaluation procedure.
\end{itemize}

\section{Related Work}
\label{sec:rel_work}
\noindent \textbf{Multi-Modal Knowledge Graph Alignment.} There exists vast literature in the domain of graph matching and ontology alignment, with methods focusing on creating single feature vectors to compare on the entire graph-level (\eg., \cite{li2019graph}). Such methods are not applicable to our case, since the two graphs to align have partial semantic and geometric overlap. A more relevant task is that of multi-modal knowledge graph (KG) alignment \cite{eva,hmea,mmkg,mkhan,mmea,cheng2022multijaf,chen2022multi,masked-mmea,mclea}, which refers to the task of aligning multiple knowledge graphs that represent information from different modalities (\eg, text, images, and videos). The goal is to integrate the knowledge from different sources and provide a more comprehensive understanding of the world. EVA \cite{eva} leverages visual and auxiliary knowledge to achieve entity alignment in both supervised and unsupervised settings, using a loss formulation inspired by NCA-based text-image matching \cite{hal}. More recently, approaches like \cite{mmea,mclea} solve the task by learning a common embedding space for all modalities, where similar entities in the KG have similar embeddings. Both approaches use multiple individual encoders to obtain modality-specific representations for each entity in the KG. However, \cite{mclea} introduces contrastive learning with intra-modal contrastive loss and inter-modal alignment loss to learn discriminative cross-modal embeddings, while \cite{mmea} only performs common space learning to align the embeddings. Since 3D scene graphs can be also considered as containing multiple modalities (\ie, object instances, their 3D geometry, attributes, and in-between relationships), we leverage the above architecture and adapt it for the setting of aligning 3D scene graphs of environments. A point to highlight is that KG alignment methods consider inputs as (partially) overlapping and true, something that in the real-world scenario of creating 3D maps does not hold due to arbitrary conditions and noise. \vspace{3pt}

\begin{figure*}[ht!]
    \centering
    \includegraphics[trim=0 0 0 0,clip,width=1.0\linewidth]{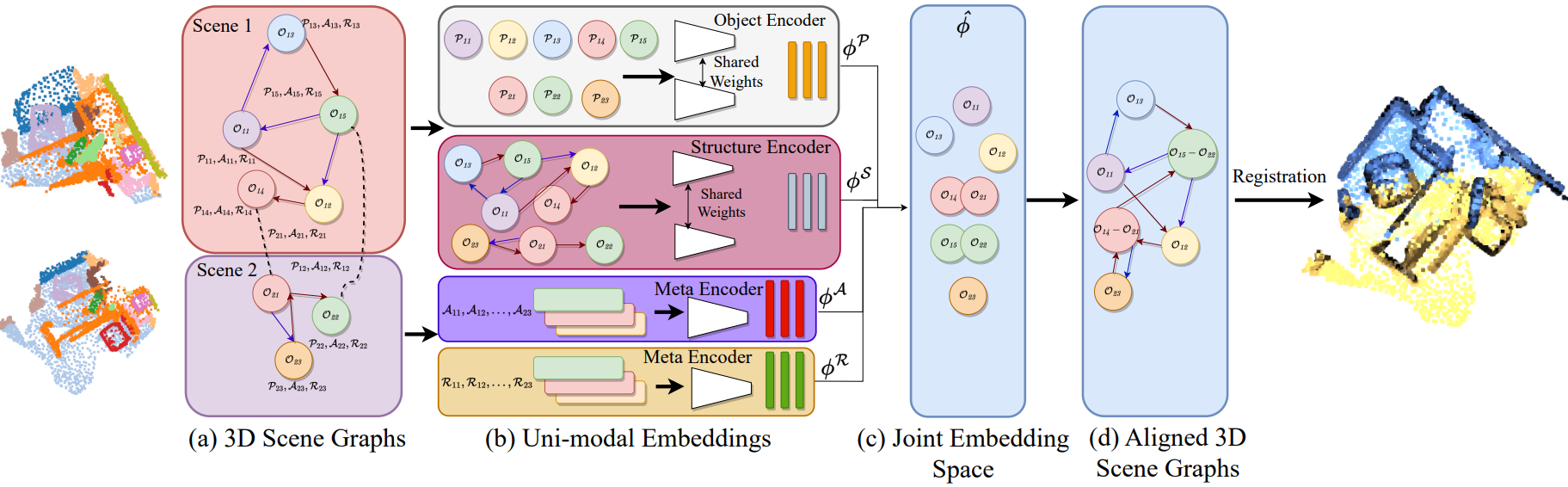}
    \caption{\textbf{Overview of \project{}.} Given two 3D scene graphs with partial overlap (a), we create four uni-modal embeddings (b) that encode object, structure, attribute, and relationship information. These interact with each other to create a joint embedding space (c) where similar nodes are located closely. We use this similarity to match nodes and create the final aligned 3D scene graph (d). We further demonstrate the use of the alignment in 3D point cloud registration.}
    \label{fig:architecture}
    \vspace{-10pt}
\end{figure*}

\noindent \textbf{3D Scene Graphs.} Following the success of utilizing 2D scene graphs \cite{krishna2017visual}, researchers introduced 3D scene graphs as structured and rich representations to describe real-world scenes \cite{armeni20193d,kim20193,rosinol20203d,3dssg}. We follow the 3D scene graph structure and dataset presented in \cite{3dssg}. Existing work addresses the task of generating 3D scene graphs with a variety of approaches; from online incremental building \cite{wu2021scenegraphfusion,hydra} to offline based on RGBD images and/or 3D reconstructions \cite{armeni20193d,3dssg,rosinol2021kimera}. Similarly, \cite{knowledgescene} advocates a knowledge-inspired scene graph prediction method based on point clouds while \cite{SGGpoint} uses edge-assisted reasoning to bridge perception and reasoning in the context of 3D scene understanding. The 3D scene graph representation has been explored in embodied AI for tasks related to navigation and planning \cite{sepulveda2018deep,rosinol2021kimera,9812179,li2022remote,dlite}, task completion \cite{Gadre_2022_CVPR,pmlr-v164-agia22a,ravichandran2020bridging,jiao2022sequential,li2022embodied}, variability estimation \cite{3dvsg}, and 3D scene manipulation \cite{dhamo2021graph}. Although prior work has demonstrated the applicability of 3D scene graphs in an increasing fashion, it has not been investigated for the task of creating 3D maps of scenes. Hydra \cite{hydra} approaches the latter using hierarchical loop closure detection and 3D scene graph optimization to complete the non-optimised scene graph, that is built on the fly, into a globally consistent and persistent one as new information is captured. On the another hand, our contribution focuses on solving graph matching solely on the object-level. This would enable to leverage, share, and recycle the created graphs for general agent operation.

\noindent \textbf{3D Point Cloud Registration.} 
The field of 3D point cloud registration is well-established and active, with approaches mainly being  feature-based and end-to-end. We focus our scope on feature-based approaches since they compute hard correspondences and perform more robustly in real-world scenes. Such methods  \cite{d3feat,predator,regtr,geotr} consist of two steps: local (learned) feature extraction and pose estimation with RANSAC \cite{martin1981random}. However, they focus on the in-vitro problem of aligning two input point clouds that have some degree of overlap. This does not always hold in a real-world scenario where there is no knowledge of whether there is any overlap or changes. 
Although some of these methods compute matchability scores per estimated 3D point correspondences (\eg, \cite{predator,geotr}), they assume overlapping input point clouds and do not have the mechanism to discard non-overlapping ones. In addition, when the number of individual point clouds to register becomes large, it requires $O(\mathcal{N}^2)$ complexity to process all possible pairs, while failing to recognize non-aligned pairs. Last, they also require large memory reserve to process input data if the point clouds are large. Our method allows processing all possible pairs faster while identifying non-overlapping pairs before performing the final registration. In addition, it is lightweight and can easily process large scenes, since, as shown in our experiments, our method can operate with a limited number of points per object instance. Please refer to Sec.\ B in the supp.\ mat for more details on this.

\section{\project{}: 3D Scene Graph Alignment}
\label{sec:method}
Following standard formulation \cite{3dssg}, we define a 3D scene graph $\mathcal{G}$ of a scene $s$ as a pair of sets $(\mathcal{N}, \mathcal{R})$ with nodes $\mathcal{N}$ and edges $\mathcal{R}$. The nodes represent 3D object instances $\mathcal{O}$ in the scan. Each node also contains a set of attributes $\mathcal{A}$ that characterizes the visual (\eg, style and texture), geometric (\eg, shape and symmetry), and state (\eg, closed and empty) characteristics of the object instance, in addition to the object categories. Instance-level point clouds $\mathcal{P}$ are available per node. The edges define semantic relationships between the nodes such as \texttt{standing on} and \texttt{attached to}. Given two graphs $\mathcal{G}_1 = (\mathcal{N}_1, \mathcal{R}_1)$ and $\mathcal{G}_2 = (\mathcal{N}_2, \mathcal{R}_2)$ of scenes $s_1$ and $s_2$ respectively, we aim to find the set of objects in the overlapping regions of the two scenes, denoted as entity pairs, $\mathcal{F} = \{(n_1, n_2) \; | \; n_1 \equiv n_2, n_1 \in N_1, n_2 \in N_2 \}$, forming node correspondences.

Our approach follows the network architecture proposed in \cite{mclea}, which we modify from the language domain for the 3D scene graph alignment task with varying degrees of overlap. 
The overall architecture is shown in Figure~\ref{fig:architecture}.
We can formulate scene graphs as a multi-modal knowledge graph -- commonly used in entity alignment -- where the semantic and geometric information included in scene graphs is treated as the different modalities that are encountered in knowledge graphs. 
The goal is to learn a joint multi-modal embedding from individual encodings of each modality (uni-modal), in which nodes are closely located if they correspond to the same underlying object instance and belong to different graphs. 
Specifically, we create uni-modal embeddings using the three main 3D scene graph information types: 
\textit{object} embedding that encodes $\mathcal{P}$, 
\textit{structure} embedding $\mathcal{S}$ that encodes $\mathcal{R}$ in the form of a structured graph, 
and two \textit{meta} modalities that encode $\mathcal{A}$ and $\mathcal{R}$ between objects in the form of a one-hot vector. 
To reason about the final entity alignment, similar to \cite{mclea}, we create a joint embedding by combining these uni-modal ones in a weighted manner and perform a joint optimization using knowledge distillation across all embeddings. 

\subsection{Uni- and Multi-Modal Embeddings}
\label{subsec:embed}

To leverage rich information in 3D scene graphs, we process each modality separately in our framework and create uni-modal embeddings, which are later processed to model complex inter-modal interactions in the joint embedding. 

\par \noindent \textbf{Object Embedding.} 
Point clouds contain rich geometric information about objects. Each of the individual point cloud $\mathcal{P}_i$ of $\mathcal{O}_i$ is the input to the object encoder. 
We employ a point cloud feature extractor backbone architecture such as \cite{qi2017pointnet, pct2021} as the object encoder to extract the visual features $\phi_{i}^\mathcal{P}$ for every node.

\par \noindent \textbf{Structure Embedding.} 
3D Scene Graphs contain information on relationships between $\mathcal{O}$, which we leverage to encode the layout of objects in $s$. 
We represent this information in the form of a \textit{structure} graph: node features are the relative translation between object instances, and edges are the aforementioned relationships. We calculate relative translation by taking the distance between the object instance consisting of the highest number of relationships and that of any other object instance in the scene. Specifically, we compute distances using the barycenter of the convex hull of the point clouds. We use a Graph Attention Network (GAT) \cite{velikovi2018graph} to model the structural information in $\mathcal{G}_1$ and $\mathcal{G}_2$ via the structure graph. 
We limit the weight matrix to a diagonal matrix, as suggested by \cite{mclea}, to minimize computations and improve the scalability of the model. As per \cite{mclea}, the neighborhood structure embedding $\phi_{i}^\mathcal{S}$ is produced by the last GAT layer, using a two-layer GAT model to aggregate the neighborhood information over several hops.

\noindent \textbf{Meta Embeddings.} 
Along with modeling the geometric and structural properties of the scene, we model the attributes and corresponding relationships per object $\mathcal{O}_i$ in two separate embeddings. We regard the relationship of $\mathcal{O}_i$ with other objects in the input scene graph as a one-hot encoded feature vector and pass it through a single-layer MLP to obtain the relational embedding $\phi_i^\mathcal{R}$. For instance, consider the set of relationships {\texttt{standing on, built in, attached to}}. If an object (node) in the input scene graph has a single edge associated with it and this represents the \texttt{built in} relationship, the input vector for the relationship encoder is $[0, 1, 0]$.  We adopt the same approach for the attributes of $\mathcal{O}_i$ for simplicity to get the $\phi_i^\mathcal{A}$.  These single-layer networks called meta encoders provide valuable insights into scene composition and facilitate straightforward extension to new data.

\noindent \textbf{Joint Embedding.} We concatenate each of the previously discussed uni-modal features to a single compact representation $\hat{\phi}_i$ for each object $\mathcal{O}_i$ as follows:
\vspace{-3pt}
\begin{equation}
    \hat{\phi}_i = \oplus_{k \in \mathcal{K}}\left[\frac{\exp(w_k)}{\sum_{j \in \mathcal{K}} \exp(w_j)} h_i^m \right],    
\end{equation}
\vspace{-3pt}

\noindent where $\oplus$ denotes concatenation, $\mathcal{K} = \{\mathcal{P}, \mathcal{S}, \mathcal{R}, \mathcal{A} \}$, and $w_m$ is a trainable attention weight for each modality $k$. 
We apply $L_2$ normalization to each uni-modal feature before the final concatenation. 
These embeddings are coarse-grained without any interaction between different modalities. 

\subsection{Contrastive Learning}
To model interaction between modalities, we formulate a representation learning framework. A contrastive loss function encourages comparable samples, or aligned entities, to be closer together and dissimilar samples to be farther away in the learned representation space. Using a cross-modal contrastive loss is a typical strategy when working with many modalities, such as in the case of 3D scene graphs, as it encourages samples from various modalities that are semantically related to be closer together in the joint representation space. Inspired by \cite{mclea}, we use the Intra-Modal Contrastive Loss (ICL) and Inter-modal Alignment Loss (IAL) and formulate them similarly.

\begin{figure}
    \centering
    \includegraphics[width=1.0\columnwidth]{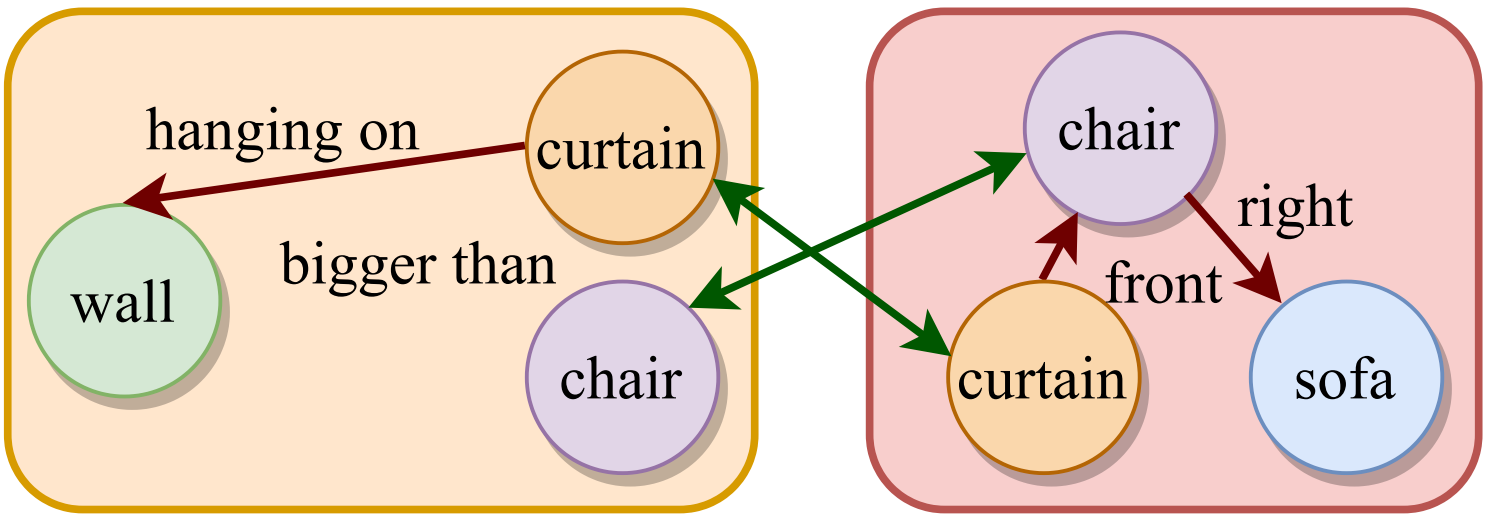}
    \caption{\textbf{An example of contrastive learning pairs.} Aligned pairs across graphs are marked with \textcolor{ForestGreen}{green} edges and unaligned pairs within the same graph are marked with \textcolor{Brown}{brown} edges.}
    \label{fig:anchor_triplet}
    \vspace{-10pt}
\end{figure}

During training, we assume that $E \subset \mathcal{F}$ is available to us as seed-aligned entity pairs. Formally, for the $i^{th}$ object  node $n_1 \in \mathcal{N}_1$, we define $E = \{ n_1^i \; | \; n_2^i \in \mathcal{N}_2 \}$, where $(n_1^i, n_2^i)$ is an aligned pair. We define the unaligned pairs within the same graph as $H_1^i = \{ n_1^j \; | \; \forall n_1^j \in \mathcal{N}_1, j \neq i \}$, and aligned pairs across graphs as $H_2^i = \{ n_2^j \; | \; \forall n_2^j \in \mathcal{N}_2, j \neq i \}$ (Figure~\ref{fig:anchor_triplet}). These two samples define the constrained joint embedding space. We model $\mathcal{L}_k^{ICL}$ to learn intra-modal dynamics for more discriminative boundaries for each modality $k$ in the embedding space. We apply ICL separately on each uni-modal embedding and on the joint concatenated embedding, after $L_2$ normalization. Each uni-modal embedding is trained individually using ICL and is not intended to interact with others.

Our complete representation is the joint embedding space, and our goal is to learn proper uni-modal encodings that enable node alignment in this joint space. To achieve this, we minimize the bi-directional KL-divergence loss between joint embedding and uni-modal embeddings as the Inter-modal Alignment Loss (IAL), thereby, emphasizing on aggregating the distribution of various modalities, which narrows the modality gap by learning interactions between various modalities inside each entity. We train our model end-to-end, optimizing both losses as follows: 
\begin{equation}
    \mathcal{L} = \mathcal{L}_o^{ICL} + \sum_{k \in \mathcal{K}} \alpha_k \mathcal{L}_k^{ICL} + \sum_{k \in \mathcal{K}} \beta_k \mathcal{L}_k^{IAL}, 
\end{equation}
where $\mathcal{K} = \{\mathcal{P}, \mathcal{S}, \mathcal{R}, \mathcal{A} \}$ and $o$ is the joint embedding. Variables $\alpha_k$ and $\beta_k$ are hyper-parameters that are automatically learned during training. 
We direct the readers to \cite{mclea} for a deeper understanding of the loss functions. 

\subsection{3D Point Cloud Registration}
\label{subsec:registration}

In this section, we describe how we approach the task of 3D point cloud registration by leveraging our scene graph alignment results. 
The output of the previously described scene alignment method is the set of matched entity pairs $n_1$ and $n_2$, in the scene graphs $\mathcal{G}_1$ and $\mathcal{G}_2$, respectively. 
For each entity pair $n_1^i$ and $n_2^i$, we extract 3D point correspondences from $\mathcal{P}_1^i$ and $\mathcal{P}_2^i$. 
The correspondences are estimated by an off-the-shelf correspondence extraction algorithm (\eg, \cite{geotr}) by running it on node pairs independently. 
We collect the point correspondences across all matched entities and then use the robust estimator, \eg RANSAC~\cite{martin1981random} or one of its recent variants~\cite{raguram2012usac,barath2018graph,barath2020magsac++}, to get the transformation $\textbf{T} \in \text{SE}(3)$ between the point clouds of the two scenes. 

Performing registration on 3D correspondences that stem from node-to-node matches allows for being less sensitive to changes in the point clouds and incorrect matching than state-of-the-art techniques. 
Such an approach has two major advantages: 
(i) it filters non-overlapping scene pairs, which should not be registered, \textbf{faster} than state-of-the-art point cloud registration methods and without any need for registration. 
This is enabled by obtaining object-level (node-to-node) correspondences instead of performing the registration directly on a large-scale 3D point cloud. 
(ii) It performs better than standard registration methods even on point clouds with low overlap, which we showcase with experiments in the following section.

\section{Experiments}
\label{sec:exps}
We evaluate \project{} on the task of 3D scene graph alignment (Section~\ref{sec:scenegraphalignment}) and on downstream applications, namely 3D point cloud registration (Section~\ref{sec:pointcloudregistration}), 3D point cloud mosaicking (Section~\ref{sec:scene_mosaick}) and 3D scene alignment with changes present in the data (Section~\ref{sec:aligningchangedscenes}). We provide additional ablation studies to further understand the performance on node matching in the supp.\ mat.

In our experiments we use the 3RScan dataset \cite{3rscan}, which consists of $1335$ indoor scenes, of which $1178$ are used for training and $157$ for testing. The dataset contains semantically annotated 3D point clouds per scene, some of which depict the same environment over time. 3D scene graph annotations for 3R Scan are provided in \cite{3dssg}. Although this allows to evaluate the robustness of our method in the case of changed environments, there are no annotations to evaluate in static environments. To enable a thorough evaluation with scenes that range in overlap (from zero to partial), we create sub-scene graphs in single scenes given their full 3D scene graph provided in the annotations, following standard point cloud registration benchmarks \cite{3dmatch}. 
The generated data is made public to maximize reproducibility. In our experiments, we use PointNet \cite{qi2017pointnet} as the object encoder, a comprehensive evaluation with multiple backbones is provided in the supp.\ mat.

\textbf{Dataset Generation.} To create the sub-scene graphs, we generate sub-scenes per scene on the geometry level. Point clouds in \cite{3rscan} are generated from RGBD frames. To imitate a realistic setting, we create sub-scenes by selecting groups of sequential frames and reconstructing the point cloud of the depicted scene. Frames across groups are unique and there is a possibility of 3D spatial overlap in the generated point clouds.   We generate a total of $6731$ sub-scenes from the training scenes and $843$ sub-scenes from the validation scenes. We create pairs using the sub-scenes generated from the same scene, such that the spatial overlap in a pair ranges from 10-90\%. This results to $15102$ pairs for training and $1932$ pairs for testing. More details, as well as statistics on the generated data are in the supp.\ mat. For simplicity, hereafter we refer to sub-scenes as \textit{scenes} ($s$ in Section~\ref{sec:method}).

The individual point clouds of object instances in the generated data will have a varying number of points. To use them as input to the object encoder (Section \ref{subsec:embed}), we require them to have the same size ($512$). We use farthest point sampling to downsample each $\mathcal{P}_i$ of object $\mathcal{O}_i$ as
\begin{equation}
    \mathcal{P}_i = \{\delta_{k^i} \odot p_k\}_{k=1, |P|},
\end{equation}
where $\delta$ represents the Kronecker delta, $p$ is a point in $\mathcal{P}$, and $|.|$ is the cardinality of $\mathcal{P_i}$, i.e, the number of points.

\begin{figure*}
    \centering
    \includegraphics[trim=0 0 0 0,clip,width=0.95\linewidth]{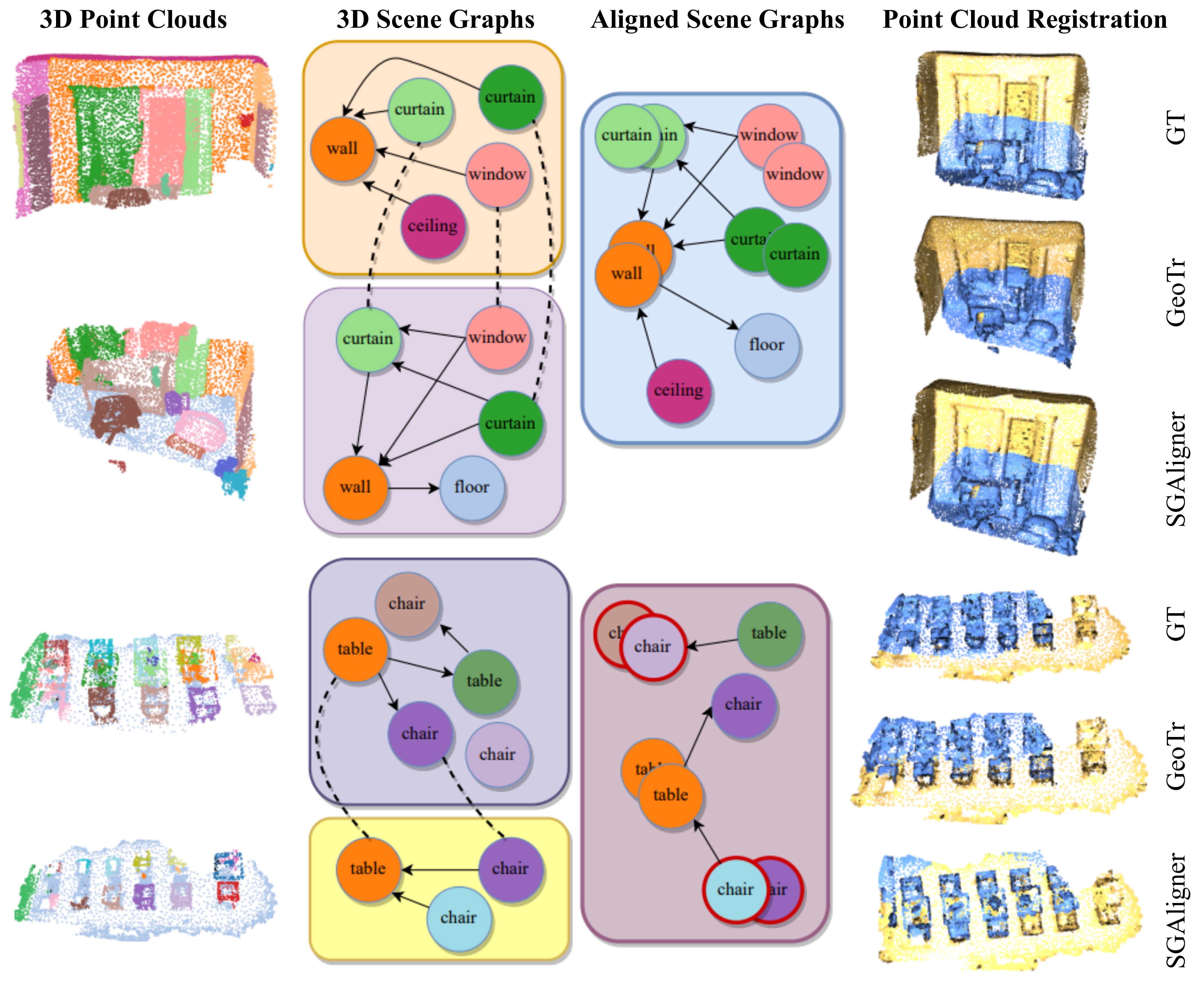}
    \caption{\textbf{Qualitative Results.} Given input 3D scene graphs (left), we showcase a success (top) and a failure (bottom) case of \project{} for alignment (middle) and registration (right). The existence of the same exact object in multiple replicas creates erroneous matches (nodes outlined in red) which cannot be recovered in registration. When objects are more discriminative, \project{} provides accurate matching.}
    \label{fig:quals}
    \vspace{-8pt}
\end{figure*}

\subsection{3D Scene Graph Alignment}
\label{sec:scenegraphalignment}
In this section, we evaluate \project{} on the 3D scene graph alignment task. We first evaluate how our method performs on node alignment. We utilize cosine similarity on the joint embedding to calculate the similarity between two matched entities and employ Mean Reciprocal Rank (MRR) and Hits@K, where K=$\{1,2,\ldots,5\}$. MRR denotes the mean reciprocal rank of correct matches. Hits@K denotes the ratio of correct matches appearing within top K, based on their cosine similarity ranking.

\begin{table}[ht!]
    \resizebox{\columnwidth}{!}{
    \centering
    \begin{tabular}{c|cccccc}
        \toprule
         \multirow{2}{*}{\textbf{Modalities}} & \textbf{Mean}  & \multicolumn{5}{c}{\textbf{Hits @ $\uparrow$}} \\
         & \textbf{RR $\uparrow$} & \textbf{K = 1} & \textbf{K = 2} & \textbf{K = 3} & \textbf{K = 4} & \textbf{K = 5} \\
        \midrule
        \multicolumn{7}{c}{\textit{\textbf{w/ Ground Truth 3D Scene Graphs}}} \\
        \midrule
        EVA \cite{eva} & 0.867 & 0.790 & 0.884 & 0.938 & 0.963 & 0.977 \\
        $\mathcal{P}$ & 0.884 & 0.835 & 0.886 & 0.921 & 0.938 & 0.951 \\
        $\mathcal{P}$ + $\mathcal{S}$   & 0.897 & 0.852 & 0.899 & 0.931 & 0.945 & 0.955 \\
        $\mathcal{P}$ + $\mathcal{S}$ + $\mathcal{R}$ & 0.911 & 0.861 & 0.916 & 0.947 & 0.961 & 0.970 \\
        \project{} & \textbf{0.950} & \textbf{0.923} & \textbf{0.957} & \textbf{0.974} & \textbf{0.982} & \textbf{0.987}\\
        \midrule
        \multicolumn{7}{c}{\textit{\textbf{w/ Predicted 3D Scene Graphs}}} \\
        \midrule
        \project{} & 0.882 & 0.833 & 0.881 & 0.918 & 0.937 & 0.951\\
        \bottomrule
    \end{tabular}
    }
    \vspace{1pt}
    \caption{\textbf{Evaluation on node matching.} We compare the performance of \project{} for different modality combinations, as well as for ground truth and predicted scene graphs.}
    \label{tab:main}
    \vspace{-10pt}
\end{table}

We compare \project{} to using different modalities, and as a result embeddings, as well as with a baseline from entity alignment in the language domain. For the latter purpose, we adapt the Entity Visual Alignment method (EVA) \cite{eva} for 3D scene graphs by replacing the visual encoder with PointNet architecture \cite{qi2017pointnet}, same as $\mathcal{P}$ in our approach. Results are in Table \ref{tab:main}. Please note that when employing only the instance level point clouds $\mathcal{P}$ there is no IAL used. As evident, our method, even when using a single modality, outperforms EVA \cite{eva} with a margin of approximately $2$\%. Furthermore, and as expected, each modality in our method contributes to improved performance in all metrics. Interestingly, using all modalities provides at K=2 better results than only $\mathcal{P}$ at k=5, and at k=3 it  is already better than any of the other combinations at k=5. To further verify the robustness of our method and hence its suitability for real-world applications, we also compare the performance of our method using both ground truth and predicted 3D scene graphs as input during inference (in both cases the network has been trained on ground truth data). We compute the scene graph predictions using the pre-trained network for 3D scene graph generation given 3D point cloud of \cite{3dssg}\footnote{ We refer the reader to Table 2 in \cite{3dssg} for an evaluation on scene graph prediction of this method.}.  \project{} is still able to provide reasonable matches despite the noise in the input, although with an expected performance drop. The success and failure case presented in Figure~\ref{fig:quals}, shows that the existence of the same exact object in multiple replicas creates erroneous matches; when the objects are more discriminative, \project{} provides accurate results. For more examples please see supp.\ mat.

We further ablate the results of our method (on ground truth data) for scenes of different overlap to evaluate the robustness in cases where few nodes can be matched. Results are reported in Table~\ref{tab:abl_overlap}. As expected, the performance drops with lower overlap, however, the gap between the very high and very low overlap is not drastic.

\begin{table}[ht!]
    \resizebox{\columnwidth}{!}{
    \centering
    \begin{tabular}{c|cccccc}
        \toprule
         \textbf{Overlap} & \textbf{Mean}  & \multicolumn{5}{c}{\textbf{Hits @ $\uparrow$}} \\
         \textbf{(\%)} & \textbf{RR $\uparrow$} & \textbf{K = 1} & \textbf{K = 2} & \textbf{K = 3} & \textbf{K = 4} & \textbf{K = 5} \\
        \midrule
        10-30 & 0.872 & 0.806 & 0.886 & 0.927 & 0.948 & 0.962 \\
        30-40 & 0.908 & 0.859 & 0.917 & 0.949 & 0.968 & 0.978 \\
        40-50 & 0.941 & 0.908 & 0.950 & 0.973 & 0.980 & 0.985 \\
        50-60 & 0.960 & 0.937 & 0.967 & 0.982 & 0.989 & 0.994 \\
        60-   & \textbf{0.979} & \textbf{0.966} & \textbf{0.982} & \textbf{0.990} & \textbf{0.994} & \textbf{0.995} \\
        \midrule
        All data & 0.950 & 0.923 & 0.957 & 0.974 & 0.982 & 0.987 \\
        \bottomrule
    \end{tabular}
    }
    \vspace{1pt}
    \caption{\textbf{Evaluation on node matching per overlap range.}}
    \label{tab:abl_overlap}
    \vspace{-8pt}
\end{table}

 In order to deeper understand the effect of semantic noise to \project{}'s performance, we provide experiments with controlled semantic noise on the test data. Specifically, we consider the following scenarios of noise: (i) only relationships are removed\footnote{Ground truth annotations do not offer an exhaustive list of relationships and attributes per node. We remove edges from these annotations.}; (ii) only object instances are removed -- their corresponding attributes and any relationships that include them are also removed; (iii) both relationships and object instances are removed; (iv) object instances are wrong (\ie they have the wrong semantic label); and (v) both relationships and objects are wrong. Noise is applied to each input scene graph randomly to 15-40\% of the modified semantic. For (iv) and (v), we randomly assign any other semantic label, \ie, a chair could be labeled as floor. Results, including those on the full ground truth dataset (GT) and with predicted 3D scene graphs (Pred.) for reference, are in Table~\ref{tab:abl_semnoise}. Note that the training set remains the same. As shown, the noise that our method can handle the best is that of missing objects (ii). This means that the relationships between objects can be more important than having structured information, however not by a large margin. What drops the performance drastically is wrong semantic labels for both object and relationship labels. Comparing the results for scenarios (iii) and (iv) with the use of predicted 3D scene graphs, we observe that for the former the values are significantly lower. This has to be taken into account in case of real-world applications, especially if generalization of the scene graph prediction algorithm is unknown.

\begin{table}[ht!]
    \resizebox{\columnwidth}{!}{
    \centering
    \begin{tabular}{cc|cccccc}
        \toprule
         & \textbf{Affected} & \textbf{Mean}  & \multicolumn{5}{c}{\textbf{Hits @ $\uparrow$}} \\
         & \textbf{Modules} & \textbf{RR $\uparrow$} & \textbf{K = 1} & \textbf{K = 2} & \textbf{K = 3} & \textbf{K = 4} & \textbf{K = 5} \\
        \midrule
        (i)   & $\mathcal{S}$, $\mathcal{R}$ & \textbf{0.906} & \textbf{0.858} & 0.915 & \textbf{0.949} & \textbf{0.964} & 0.974 \\
        (ii)  & All & \textbf{\underline{0.924}} & \textbf{\underline{0.878}} & \textbf{\underline{0.942}} & \textbf{\underline{0.968}} & \textbf{\underline{0.977}} & \textbf{\underline{0.985}} \\
        (iii) & All & 0.902 & 0.848 & \textbf{0.918 }& \textbf{0.949} & \textbf{0.964} & \textbf{0.975} \\
        (iv)  & $\mathcal{P}$ & 0.661 & 0.563 & 0.643 & 0.699 & 0.743 & 0.776 \\
        (v)   & $\mathcal{P}$, $\mathcal{S}$, $\mathcal{R}$ & 0.587 & 0.479 & 0.573 & 0.632 & 0.674 & 0.709 \\ 
        \midrule
        GT & None & 0.950 & 0.923 & 0.957 & 0.974 & 0.982 & 0.987 \\
        Pred. & None & 0.882 & 0.833 & 0.881 & 0.918 & 0.937 & 0.951 \\
        \bottomrule
    \end{tabular}
    }
    \vspace{1pt}
    \caption{\textbf{Evaluation on node matching with controlled semantic noise.} \textit{Best} values are in \underline{\textbf{underlined bold}}. \textit{Second best} in \textbf{bold}.}
    \vspace{-5pt}
    \label{tab:abl_semnoise}
\end{table}

\noindent \textbf{3D Scene Graph Alignment.} Entity alignment provides pairs of matched nodes. Here, we evaluate how well two scene graphs can be aligned given the predicted node matching. In theory, since the scenes are rigid, two matched nodes would be enough to perform the alignment, if there was no noise in the matches. Since in reality matches are noisy, we evaluate 3D scene graph alignment with K equals top-2, top-50\%, and all of the matched nodes. To measure this performance we introduce the \textit{Scene Graph Alignment Recall (SGAR)} metric. Similar to the standard recall metric, we calculate the amount of correct alignments of \project{}. We provide these results in Table~\ref{tab:scene_alignment_recall} for both ground truth and predicted scene graphs. As shown, results with K=2 perform the best in both cases. This means that \project{} can identify at least two matches that are very closely located in the joint embedding space. In addition, SGAR for the top-2 matches is approximately the same for both ground truth and predicted scene graphs, which further shows that our approach can robustly align them. 

\begin{table}[ht!]
    \centering
    \resizebox{.88\columnwidth}{!}{
    \begin{tabular}{cc|ccc}
        \toprule
        \multicolumn{2}{c|}{\textbf{Input}}  & \multicolumn{3}{c}{\textbf{Scene Graph Alignment Recall} $\uparrow$} \\
         \multicolumn{2}{c|}{\textbf{Scene Graphs}} & \textbf{R@top-2} & \textbf{R@top-50\%} & \textbf{R@All} \\
         \midrule
         \multicolumn{2}{c|}{Ground Truth} & \textbf{0.964} & 0.948 & 0.738 \\
         \multicolumn{2}{c|}{Predicted} & \textbf{0.963} & 0.856 & 0.450 \\
        \bottomrule
    \end{tabular}
    } 
    \vspace{5pt}
    \caption{\textbf{Evaluation on 3D scene graph alignment.} We report for both ground truth and predicted scene graphs.}
    \label{tab:scene_alignment_recall}
    \vspace{-10pt}
\end{table}

\subsection{3D Point Cloud Registration}
\label{sec:pointcloudregistration}
In this section, we evaluate the performance of \project{} for 3D point cloud registration on the same data. We employ the state-of-the-art Geotransformer \cite{geotr} as a 3D correspondence extraction method, as per Section~\ref{subsec:registration}. We compare the performance of our approach to standard approaches and use Geotransformer directly on the point clouds of the two scenes to register. Please note that in our case, we use Geotransformer to extract correspondences from individual point clouds on the level of object instances only for the matched nodes. In all cases, we use the Geotransformer model trained on the 3DMatch dataset \cite{3dmatch}.

\textbf{Metrics:} We compute the standard metrics of Chamfer distance (CD) as in \cite{rpmnet}, relative rotation error (RRE), relative translation error (RTE), feature match recall (FMR), and registration recall (RR). RR is calculated with the standard threshold of $\text{RMSE}=0.2$.

\begin{figure}
\centering
\includegraphics[width=\columnwidth]{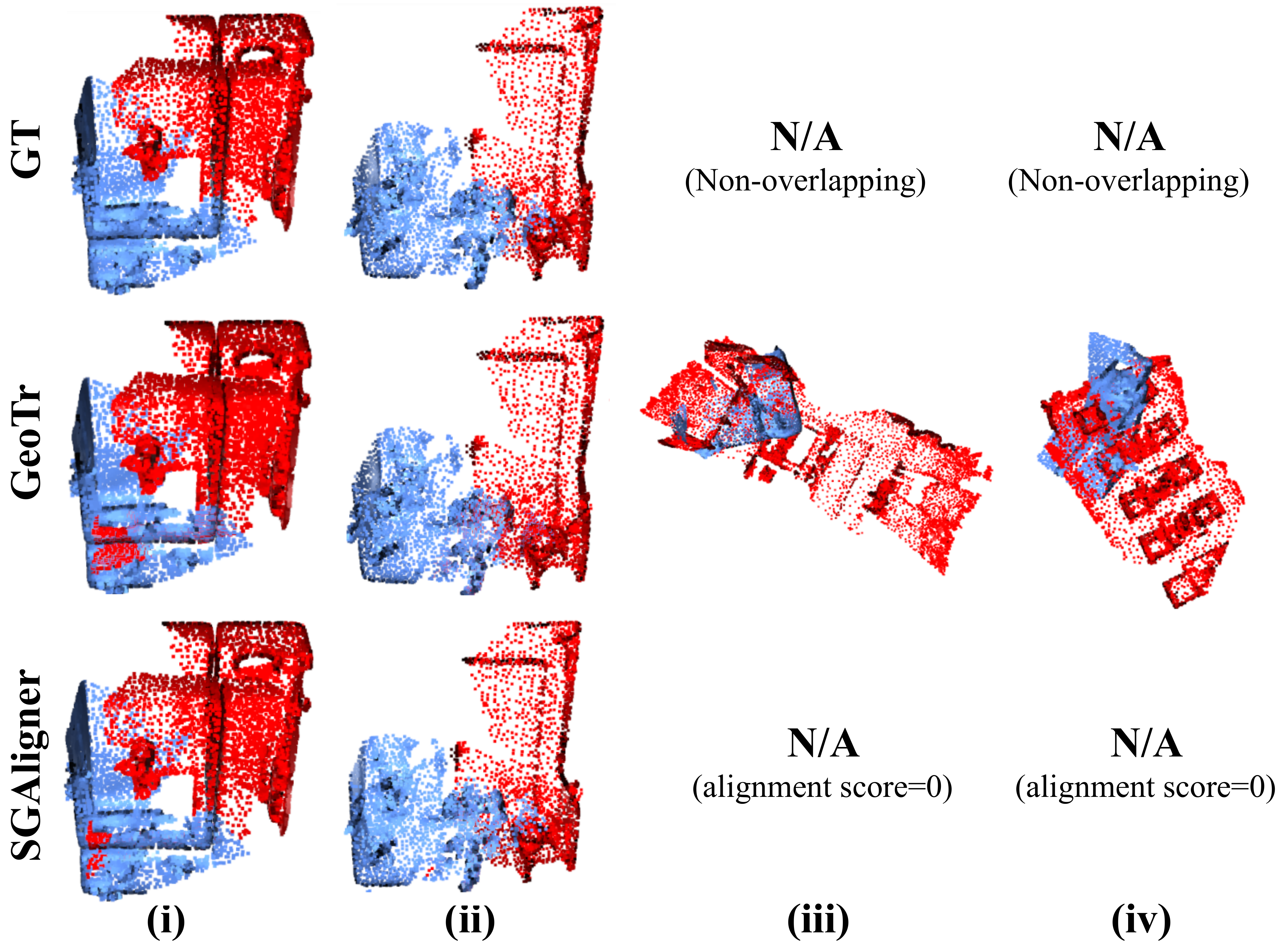}
\caption{\textbf{Registration Results for \project{} and \cite{geotr}}. Columns (i)-(ii) represent registration on overlapping pairs. (iii)-(iv) show incorrect results of \cite{geotr} for \textbf{non-overlapping} pairs, where our \textit{alignment score} is zero (\ie, there is no registration output to report for \project{}). Since these pairs are not overlapping, there is also no ground truth registration.}

\label{fig:registration}
\vspace{-15pt}
\end{figure}

Results for 3D point cloud registration for \textit{overlapping scenes} are shown in Table~\ref{tab:registration}. When using predicted data from
SceneGraphFusion \cite{wu2021scenegraphfusion}, our method not only remains robust but
also shows even higher gains, as we leverage node-to-node alignment for improved local context
compared to global registration on noisy point cloud predictions. Our method is consistently providing best results in all metrics w.r.t. the standard registration approach, despite the less geometric information in the point clouds we use; specifically on CD, we provide a 49.4\% improvement (K=2). This is an expected behavior since our alignment method, even when it contains incorrect matches, is still providing an initialization to the task and narrows down the search space for correspondences. With respect to self baselines, our method with K=2 performs the best. The drop in K=3 can be attributed to the following: more matches per each node of which one is correct, means that more outliers are provided to RANSAC which it cannot easily remove. Since the gap of Hits@K between K=2 and K=1 is larger than that between K=3 and K=2, K=2 can still provide a boost of inliers to RANSAC even though it will also increase outliers with respect to K=1. We include examples on in Figures~\ref{fig:quals} and \ref{fig:registration}.

\begin{table}[ht!]
    \resizebox{\columnwidth}{!}{
    \centering
    \begin{tabular}{cc|ccccc}
        \toprule
        \multicolumn{2}{c|}{\textbf{Methods}}  & \textbf{CD} $\downarrow$ & \textbf{RRE} ($^\circ$) $\downarrow$  & \textbf{RTE} (cm) $\downarrow$ & \textbf{FMR} (\%) $\uparrow$ & \textbf{RR}(\%) $\uparrow$ \\
        \midrule
        \multicolumn{6}{c}{\textit{\textbf{w/ Ground Truth 3D Scene Graphs}}} \\
        \hline
        \multicolumn{2}{c|}{GeoTr} & 0.02247	& 1.813 & 2.79 & \textbf{98.94} & 98.49 \\
        \parbox[t]{2mm}{\multirow{3}{*}{\rotatebox[origin=c]{90}{\textbf{Ours}}}} & K=1 & 0.01677 & \textbf{1.425} & 2.88 & \underline{\textbf{99.85}} & 98.79 \\
        & K=2 & \underline{\textbf{0.01111}} & \underline{\textbf{1.012}} & \underline{\textbf{1.67}} & \underline{\textbf{99.85}} & \underline{\textbf{99.40}} \\
        & K=3 & \textbf{0.01525}	& 1.736 & \textbf{2.55}	& \underline{\textbf{99.85}} & \textbf{98.81} \\
        \midrule
        \multicolumn{6}{c}{\textit{\textbf{w/ Predicted 3D Scene Graphs}}} \\
        \hline
        \multicolumn{2}{c|}{GeoTr} & 0.06643 & 5.697  & 9.54 & 92.23 & 93.15 \\
        \parbox[t]{2mm}{\multirow{3}{*}{\rotatebox[origin=c]{90}{\textbf{Ours}}}} & K=1 & 0.05041 & 2.49 & 3.86 & 95.25 & 94.95 \\
        & K=2 & 0.04251 & 1.725 & 3.36 & 97.12 & 98.33 \\
        & K=3 & 0.04863	& 2.194 & 2.55	& 96.83 & 97.96 \\
        
        \bottomrule
    \end{tabular}
    } \vspace{1pt}
    \caption{\textbf{3D Point Cloud Registration.} \textit{Best} values are in \underline{\textbf{underlined bold}}. \textit{Second best} in \textbf{bold}.}
    \label{tab:registration}
\end{table}


We further ablate the results on this task for scenes of different overlap. As shown in Table \ref{tab:registration_peroverlap}, our approach outperforms the standard one per overlap, and even provides better results for scenes with 30\% overlap or higher than the standard approach can do on 60\% or higher.

\begin{table}[ht!]
    \resizebox{\columnwidth}{!}{
    \centering
    \begin{tabular}{c|c|ccccc}
        \toprule
         \multicolumn{2}{c}{\textbf{Overlap}} (\%) & \textbf{CD} $\downarrow$ & \textbf{RRE} $\downarrow$ ($^\circ$) & \textbf{RTE} (cm) $\downarrow$ & \textbf{FMR} (\%) $\uparrow$ & \textbf{RR} (\%) $\uparrow$ \\
        \midrule
        \parbox[t]{2mm}{\multirow{3}{*}{\rotatebox[origin=c]{90}{\textbf{GeoTr.}}}}
                                             & 10-30 & 0.09788 & 8.830 & 13.56 & 94.57 & 92.25 \\
                                             & 30-60 & 0.00584 & 0.156 & 0.24  & 97.28 & 97.36 \\
                                             & 60-   & 0.00177 & 0.048 & 0.07  & 99.47 & 99.31 \\
        \midrule
        \parbox[t]{2mm}{\multirow{3}{*}{\rotatebox[origin=c]{90}{\textbf{Ours}}}}
                                             & 10-30 & 0.05160 & 5.660 & 8.48 & 99.23 & 95.35 \\
                                             & 30-60 & \textbf{0.00127} & \textbf{0.045} & \textbf{0.05} & \textbf{99.68} & \textbf{98.34} \\
                                             & 60-   & \underline{\textbf{0.00046}} & \underline{\textbf{0.018}} & \underline{\textbf{0.02}} & \underline{\textbf{99.92}} & \underline{\textbf{99.93}} \\
        \bottomrule
    \end{tabular}
    }
    \vspace{1pt}
    \caption{\textbf{3D Point Cloud Registration per overlap.} For our method we use the best performing (K=2). \textit{Best} values are in \underline{\textbf{underlined bold}}. \textit{Second best} in \textbf{bold}.}
    \label{tab:registration_peroverlap}
\end{table}

\noindent \textbf{Overlapping vs Non-Overlapping Scenes.} 
In practical and real-world applications, we do not always know if two scenes are overlapping or not. While standard point cloud registration approaches compute a matchability score, they typically train and test only on \textit{overlapping} scenes and do not have the mechanism to discard non-overlapping ones.
Here, we demonstrate how our approach can indirectly provide a solution to this problem, without any additional supervision. 
We formulate it as to how well a method can identify whether a pair of scenes overlaps. 
For the state-of-the-art Geotransformer, we compute a scene-level \textit{matchability} $\mathcal{\mu}$ by averaging over all correspondence matchability scores and consider two scenes as overlapping if $\mathcal{\mu} \ge 0.2$. 
For our approach, we compute a scene-level \textit{alignment} score $\mathcal{\xi}$ by averaging the total number of matched nodes (for K=1) and consider two scenes as overlapping if $\mathcal{\xi} \ge 0.2$.

To evaluate this task, we create a new test set that includes overlapping and non-overlapping scenes from the data we generate. Specifically, we use $1932$ of the overlapping pairs from the former, and sample from the rest of the scenes $1932$ non-overlapping pairs. Please note that the training set remains the same and contains \textit{only} overlapping pairs. We compute the precision, recall, F1-score, and time in milliseconds required to make the overlap decision for all pairs. Results are in Table \ref{tab:non_overlapping}. Our approach runs 3 times faster than Geotransfomer since it does not process the entire point cloud, which can be computationally demanding. In addition, it leads to comparative performance while being able to identify more overlapping pairs correctly. 

We choose $\mathcal{\mu}$ and $\mathcal{\xi}$ based on the best performing value of $\mathcal{\mu}$ for Geotransformer. Specifically, we observed that a higher value (0.4) leads to no true positives and a lower one (0.1) leads to lower precision (66.94\%) since it identifies more false positives. 
This is understandable given that it was not trained on this data. 
However, (0.2) for a metric measuring average point correspondence similarity is low. 
For $\mathcal{\xi}$ we observe only a slight decrease in precision instead of the best performing value (0.4). In contrast, our approach does not require such difficult-to-set thresholding scheme.

\begin{table}[ht!]
    \resizebox{\columnwidth}{!}{
    \centering
    \begin{tabular}{c|ccccc}
        \toprule
        \textbf{Method} & \textbf{Prec.} (\%) $\uparrow$ & \textbf{Recall} (\%) $\uparrow$ & \textbf{F1} (\%) $\uparrow$ & \makecell{\textbf{Average Time} \\ \textbf{Per Scene} (ms) $\downarrow$} \\
        \midrule
        Geotr & \textbf{99.63} & 80.98	& 89.34 & 442.50 \\
        Ours  & 92.03 & \textbf{90.94}	& \textbf{91.48} & \phantom{1}\textbf{139.64} \\
        \bottomrule
        \end{tabular}
    }
    \vspace{1pt}
    \caption{\textbf{Overlap Check for Point Cloud Registration.} }
    \label{tab:non_overlapping}
    \vspace{-10pt}
\end{table}

\subsection{3D Point Cloud Mosaicking}
\label{sec:scene_mosaick}
In this section, we aim at reconstructing the 3D scene from a set of partial point clouds observing parts of the scene. 
We proceed by selecting one of the point clouds as the origin and then estimating the absolute pose (\ie, 3D translation and rotation) between the origin and each remaining point cloud in the set.
One can imagine this procedure as 3D point cloud mosaicking. 
Please note that there is no guarantee that all partial clouds overlap with the one chosen as origin. 
While this could be alleviated by forming all possible pairs and forcing global consistency, solving the 3D mosaicking problem falls outside the scope of this paper. 
We only aim to demonstrate the potential of the proposed algorithm for this problem.

We perform the pairwise registration for all pairs using the method described in Section \ref{subsec:registration}. In Table \ref{tab:reconstruction}, we report results and compare with Geotransformer \cite{geotr}, using the same reconstruction paradigm. The evaluation metrics we use focus on the geometric aspects of accuracy and completeness \cite{sun2021neucon} of the resulting reconstruction, as well as on precision, recall, and F1-score of registered point clouds. \project{} has higher performance on 3 out of 5 metrics, and is mainly affected in completion and precision. The performance drop is due to the fact that for some scenes with low overlap, \project{} fails to perform node alignment and, hence, registration with the incorrect alignments fails. In these scenes, GeoTr has a better complete context of the entire scene unlike the only object-based context in our approach. Furthermore, it is interesting to note that our K=1 method performs better than K=2. This is expected since there are more spurious matches from K=2 that lead to worse performance. We show qualitative results of success and failure in supp.\ mat.

\begin{table}[h]
    \resizebox{\columnwidth}{!}{
    \centering
    \begin{tabular}{cc|ccccc}
        \toprule
        \multicolumn{2}{c|}{\textbf{Methods}}  & \textbf{Acc} $\downarrow$ & \textbf{Comp} $\downarrow$  & \textbf{Precision} $\uparrow$ & \textbf{Recall} $\uparrow$ & \textbf{F1-Score}  $\uparrow$ \\
        \midrule        
        \multicolumn{2}{c|}{GeoTr \cite{geotr}} & 0.20721 & \textbf{0.03835} & \textbf{0.94180} & 0.79118 & 0.83667 \\
        
        \parbox[t]{2mm}{\multirow{2}{*}{\rotatebox[origin=c]{90}{\textbf{Ours}}}} & K=1 &  \textbf{0.00944} & 0.09347 & 0.90873 & \textbf{0.97444} & \textbf{0.93575} \\
        & K=2 & 0.01215 & 0.10641 & 0.89234 & 0.97042 & 0.92345\\
        \bottomrule
    \end{tabular}
    } \vspace{1pt}
    \caption{\textbf{Point cloud mosaicking from multiple point clouds.} \project{} performs best in 3 out of the 5 metrics, with K = 1. The drop in the other 2 is due to a few low overlap pairs where node alignment fails, hence, registration too. \textit{Best} values in \textbf{bold}.}
    \label{tab:reconstruction}
\end{table}

\subsection{Aligning 3D Scenes with Changes}
\label{sec:aligningchangedscenes}

In this section, we investigate the task of aligning a new 3D scene (target) on a prior 3D map (source), where the new scene can overlap fully or partially with the prior map and may contain changes (both geometric and semantic). Specifically, we investigate the following scenarios: (i) aligning a \textit{local} 3D scene on a \textit{larger} prior map that contains no changes -- here overlap of the local scene with the map is 100\%; (ii) aligning a 3D scene on a prior map that contains changes; and (iii) aligning a \textit{local} 3D scene on a \textit{local} prior map that contains changes. We approach this as a 3D scene graph alignment task. For (i) and (ii), we use as large maps the 3D scene graphs of the entire scenes offered in \cite{3rscan,3dssg}. For local 3D scenes we use the data generated above. The results in Table~\ref{tab:localization} show that performance depends on the size of the prior map, whether there is full or partial overlap, and on the existence of temporal differences. They also demonstrate that our method can handle the alignment of maps that showcase temporal changes, even if not explicitly trained for this purpose.

\begin{table}[ht!]
    \resizebox{\columnwidth}{!}{
    \centering
    \begin{tabular}{c|cccccc|c}
        \toprule
        & \textbf{Mean}  & \multicolumn{5}{c|}{\textbf{Hits @ $\uparrow$}} & \textbf{No. of}\\
        & \textbf{RR $\uparrow$} & \textbf{K = 1} & \textbf{K = 2} & \textbf{K = 3} & \textbf{K = 4} & \textbf{K = 5} & \textbf{Pairs}\\
        \midrule
        (i)   & 0.970 & 0.952 & 0.976 & 0.989 & 0.993 & 0.995 & 827 \\
        (ii)  & 0.934 & 0.907 & 0.933 & 0.960 & 0.966 & 0.972 & 110 \\
        (iii) & 0.886 & 0.833 & 0.894 & 0.928 & 0.946 & 0.957 & 2262 \\
        \bottomrule
    \end{tabular}
    }
    \vspace{1pt}
    \caption{\textbf{Alignment of a local 3D scene to a prior 3D map with differences in overlap and changes.}}
    \label{tab:localization}
    \vspace{-15pt}
\end{table}

\section{Conclusion}
\label{method}

We presented SGAligner, the first method capable of aligning 3D scene graphs directly on the graph level, that is robust to the in-the-wild scenarios, such as unknown overlap between scenes or changing environments. 
We demonstrated that aligning the scenes directly on the scene graph level can improve downstream tasks (\eg, point cloud alignment) in terms of accuracy and speed.    
We believe our work could unlock agents to leverage this emerging scene representation for creating 3D maps of the environment, further using it for and sharing it with downstream tasks.

\section{Acknowledgement}
D. Barath was supported by the ETH Postdoc Fellowship and the Hasler Foundation.

\renewcommand{\thesection}{\Alph{section}}

\title{\textbf{Supplementary Material \\ \project{}: 3D Scene Alignment with Scene Graphs}
\vspace{-3.5em}}
\author{}
\date{}

\maketitle
\thispagestyle{empty}
\setcounter{section}{0}

\begin{abstract}
In the supplemental material, we provide additional details about the following:
\begin{enumerate}[leftmargin=12pt,itemsep=0em]
    \item Visualisation on point cloud mosaicking given multiple individual observations (Section \ref{sec:scene_reconstruct}),
    \item Comparison to a retrieval-based approach (Section \ref{sec:baseline_comparison}),
    \item Additional ablation on SGAligner to further understand the performance of node matching (Section \ref{sec:self_ablation}),
    \item Information on the SGAligner benchmark, including details on the generated data, evaluation protocol, and metrics (Section \ref{sec:benchmark}), and
    \item Details on implementation (Section \ref{sec:implementation}).
\end{enumerate}
\end{abstract}
\vspace{-6pt}

\section{Application: Point Cloud Mosaicking}
\label{sec:scene_reconstruct}
\begin{figure*}[tp]
    \centering
    \includegraphics[trim=0 0 0 5,clip,width=0.97\linewidth]{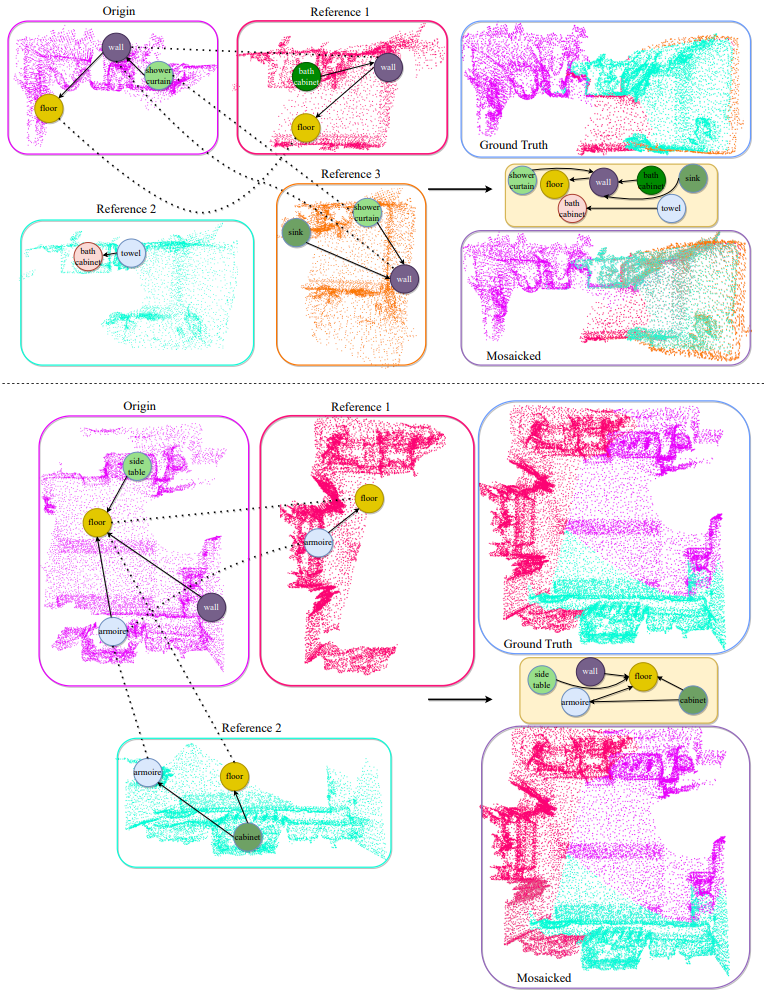}
    \caption{\textbf{Qualitative Results on Point Cloud Mosaicking.} Given partial point clouds of a scene and the corresponding 3D scene graphs, we showcase two example results on Point Cloud Mosaicking and the creation of a unified scene graph using the methodology discussed in Section \ref{sec:scene_reconstruct}. The solid lines show the relationships between objects and dashed lines represent the ground truth entity pairs $\mathcal{F}$.}
    \label{fig:scene_reconstruction}
\end{figure*}

\begin{figure*}[tp]
    \centering
    \includegraphics[trim=0 0 0 7,clip,width=\linewidth]{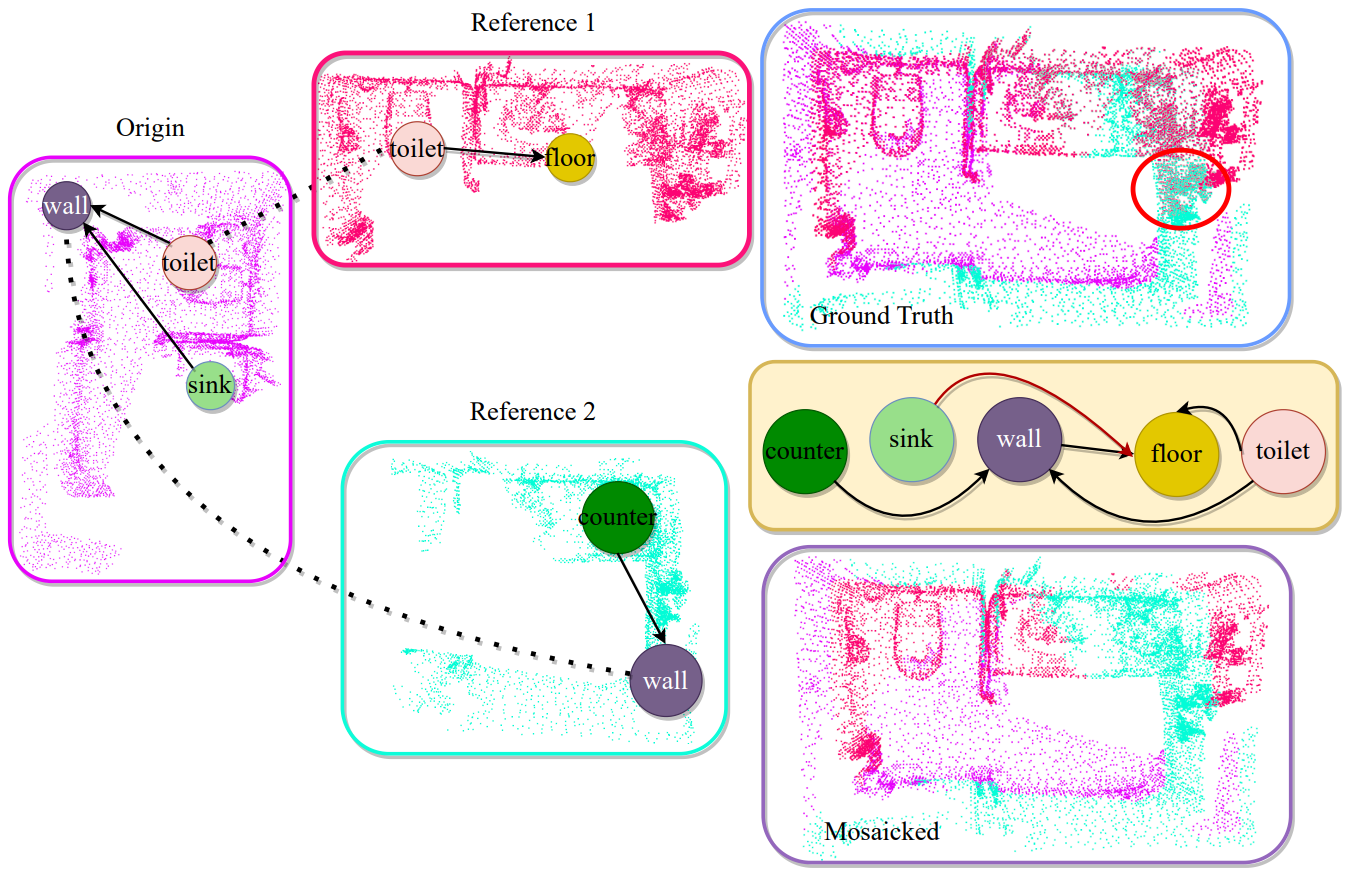}
    \caption{\textbf{Failure on Point Cloud Mosaicking.} Similar to the success cases in
    Fig. \ref{fig:scene_reconstruction}, we also showcase a failure case of our approach on point cloud mosaicking. The solid lines show the relationships between objects and dashed lines represent the ground truth entity pairs $\mathcal{F}$. The  red circle on the ground truth point cloud depicts the area where the failure is the most visible and red arrow demonstrates the misalignment of \project{} between a \textit{sink} and a \textit{floor}.}
    \label{fig:mosaick_fail}
\end{figure*}

In Section 4.3 of the main paper, we demonstrate the potential of \project{} on 3D point cloud mosaicking. Here, two success cases are shown in Figure \ref{fig:scene_reconstruction} and a failure in Figure \ref{fig:mosaick_fail} (the graphs are shown simplified for visualisation purposes and do not represent the entire available graph).

\section{Application: Finding Overlapping Scenes}
\label{sec:baseline_comparison}
In the main paper, we discussed that \project{} provides less than $O(N^2)$ computation complexity when addressing the task of registering multiple 3D scenes for which we have no knowledge of whether they overlap or not. Another approach to avoid full registration on all pairs (standard registration methods), is to use a retrieval-based approach. We consider the following approach as baseline: (i) extract local 3D keypoints for all available 3D point clouds \cite{iss}; (ii) generate a 3D descriptor per extracted keypoint \cite{fpfh}; (iii) accumulate the 3D keypoint descriptors into a global descriptor for each point cloud \cite{vlad}, and (iii) perform kNN search to rank global descriptors based on the queried one. Similarly here, this experiment serves as a demonstration of the potential of \project{} and does not aim to solve the task.

Specifically, given a point cloud, we extract keypoints from the entire scene based purely on geometry and without any notion of object-ness or semantics. We randomly select 500 keypoints and their descriptors per scene, which we use to train \cite{vlad} so as to generate optimized global descriptors. During inference and given a query point cloud and its corresponding global descriptor, we perform a kNN search to identify the closest neighbors of it in the rest of the point clouds. We evaluate on Mean Reciprocal Rank (MRR) and compare with \project{}. 

Results are shown in Table \ref{tab:vlad_subsample}. We evaluate on different point cloud densities, ranging from using the full point cloud density offered in \cite{3rscan} to random subsampling for $10$, $20$, $30$, and $50$\%. Please note that in \project{}, we do not use the entire scene, only objects in the scene graph. As described in the main paper, we downsample object point clouds using farthest point sampling to $512$ points. We follow the same protocol here and perform this operation per subsampled scene level.

As expected, the retrieval-based method is performing well when there is a dense point cloud, since our method employs a limited amount of points per object instance. However, VLAD+KNN cannot retain a robust performance when density decreases, already reaching lower performance than \project{} at 10\% subsampling. In contrast, our approach is barely affected by a changing density since it already operates on lower-resolution point clouds. This showcases that the topological information encoded in 3D scene graphs can lead to more robust results when dealing with common failure cases in global descriptors (\ie, changes in point cloud density).

\begin{table}[ht!]
    \centering
    \begin{tabular}{c|cc}
        \toprule
        \textbf{Subsampling \%} & \textbf{VLAD + KNN} & \textbf{\project{}}  \\
        \midrule
        0  & \underline{\textbf{0.557}} & 0.383  \\
        10 & 0.316	& \textbf{0.356} \\
        20 & 0.276	& \textbf{0.343} \\
        30 & 0.222	& \textbf{0.339} \\
        50 & 0.162  & \textbf{0.312} \\
        \bottomrule
        \end{tabular}
    \vspace{2pt}
    \caption{\textbf{Mean Reciprocal Rank ($\uparrow$) comparison with a retrieval-based approach.} \textit{Best} results per subsampling level are in \textbf{bold}. \textit{Overall best} in \textbf{\underline{underlined bold}}.}
    \label{tab:vlad_subsample}
\end{table}

\section{Additional Ablation Studies}
\label{sec:self_ablation}

\subsection{Analysis with Various Object Encoders}

In our experiments, we choose PointNet \cite{qi2017pointnet} as our encoder because it is commonly employed in most scene graph methods \cite{wu2021scenegraphfusion}, \cite{SGGpoint}, \cite{knowledgescene}. Pointnet has been shown to perform in real-time scenarios \cite{realtimepointnet}, which makes it suitable for mobile robot applications. In Table \ref{tab:comparison}, we present a comparison of object encoders, on the node matching task. Point Cloud Transformer (PCT) \cite{pct2021}, being inherently permutation invariant to an unordered point cloud, shows an improvement in the metrics. These results also showcase that our method is robust and agnostic to the 3D visual encoder.

\begin{table}[ht!]
    \vspace{-6pt}
    \resizebox{\columnwidth}{!}{
    \centering
    \begin{tabular}{c|cccccc}
        \hline
         \multirow{2}{*}{\textbf{Encoder}} & \textbf{Mean}  & \multicolumn{5}{c}{\textbf{Hits @ $\uparrow$}} \\
         & \textbf{RR $\uparrow$} & \textbf{K = 1} & \textbf{K = 2} & \textbf{K = 3} & \textbf{K = 4} & \textbf{K = 5} \\
        \hline
        PointNet \cite{qi2017pointnet} & 0.950 & 0.923 & 0.957 & 0.974 & 0.982 & 0.987 \\
        PCT \cite{pct2021} &  \textbf{0.965} & \textbf{0.947} & \textbf{0.968} & \textbf{0.983} & \textbf{0.988} & \textbf{0.991} \\
        \hline
    \end{tabular}
    }
    \vspace{-2pt}
    \caption{\textbf{Comparison on node matching of \project{} using various object encoders.} Best values are in \textbf{bold}.}
    \label{tab:comparison}
    \vspace{-5pt}
\end{table}

\subsection{Intra-Graph Alignment Recall}
To further validate how our model performs on aligning nodes between two 3D scene graphs (source-reference) with no/partial overlap, we formulate \textit{Intra-Graph Alignment Recall (IGAR)} metric. It measures what fraction of the nodes in the source graph are aligned (K=$1$), with nodes in the same source graph or, in other words, how many node matches out of total are self-aligned. We provide these results in Table \ref{tab:self_align}. We do not explicitly model \textbf{not} self-matching nodes within the same graph, yet, \textit{IGAR} values stand to show that our method rarely performs this.

\begin{equation}
    IGAR = \frac{1}{M} \sum_{M} \frac{| pred\{n^i \equiv n^j\} |}{ |\mathcal{F}|},  n \in \mathcal{N}
\end{equation}

where, $i \ne j$, $pred\{n^i \equiv n^j\}$ is the set of nodes in the graph which \project{} aligned with nodes in the same graph, $\mathcal{F}$ is the set of ground truth anchor pairs and $\mathcal{N}$ denotes the set of objects in a single graph and $M$ is the total number of graphs. 
 
\begin{table}[ht!]
    \centering
    \begin{tabular}{c|c}
        \toprule
        \textbf{Method} & \textbf{IGAR} $\downarrow$ (\%) \\
        \midrule
        $\mathcal{P}$ & 16.9 \\
        $\mathcal{P}$ + $\mathcal{S}$  & 16.5  \\
        $\mathcal{P}$ + $\mathcal{S}$ + $\mathcal{R}$ & 13.1 \\
        $\project{}$ & \textbf{\phantom{1}{8.2}} \\
        \bottomrule
        \end{tabular}
    \vspace{1pt}
    \caption{\textbf{Evaluation on node self-alignment.} \project{} has not been explicitly modeled to not create self-matches but still is able to differentiate between nodes from the same and different graphs.}
    \label{tab:self_align}
\end{table}

\begin{figure*}[ht!]
    \centering
    \includegraphics[trim=0 0 0 0,clip,width=1.03\linewidth]{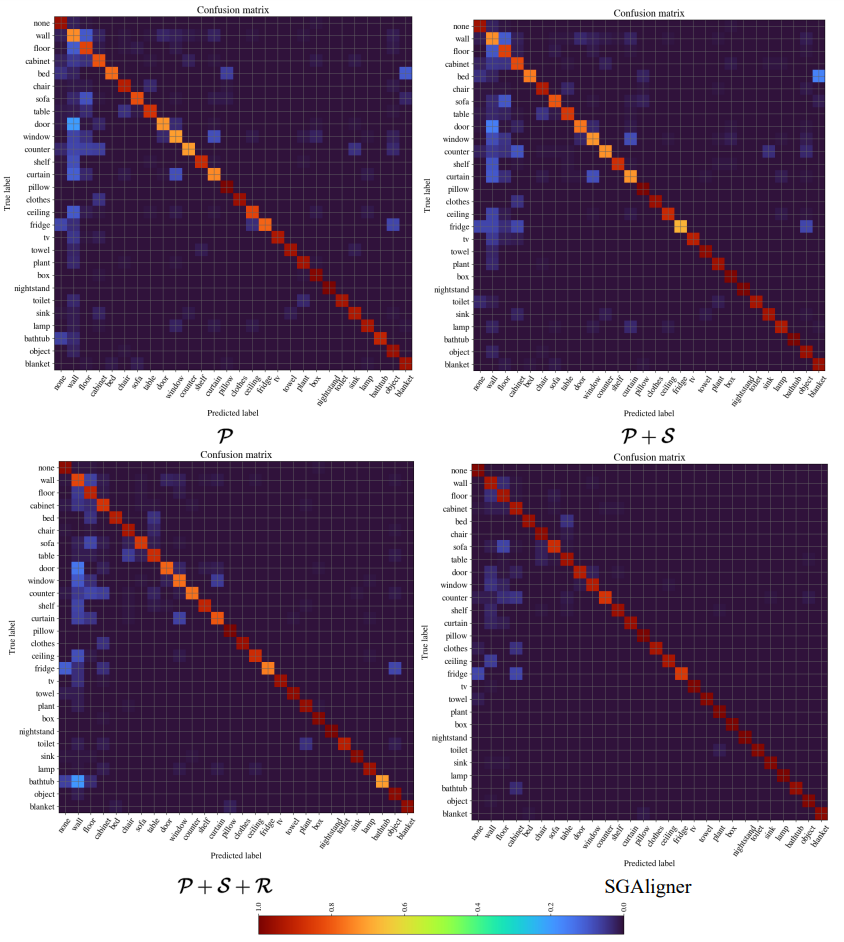}
    \caption{\textbf{Object Confusion Matrices} of the $4$ module combinations of \project{}: object encoder ($\mathcal{P}$), object and structure encoders ($\mathcal{P} + \mathcal{S}$), object, structure and relationship encoders ($\mathcal{P} + \mathcal{S} + \mathcal{R}$), and the proposed method with all modules (\project{}).
    High values indicate that an object (denoted on $y$-axis) is often recognized as the object denoted on $x$-axis -- everything but the diagonal should be 0.
    }
    \label{fig:confusion_mat}
\end{figure*}
\subsection{Confusion Matrix}
We compute a confusion matrix to identify which object categories are most frequently misaligned during entity alignment and if our method fails on certain semantic classes (\eg, \textit{chair}, \textit{table}, etc). In Figure \ref{fig:confusion_mat}, we show the confusion matrix on all $4$ module combinations of \project{}. As expected, the object encoder module $\mathcal{P}$, although performing well, confuses the most the \textit{wall} and \textit{floor} classes. This is due to the fact that purely on a semantic level, without encoding any positional/structural information, these classes are similar. We can further observe that \project{} is robust to certain classes like \textit{pillow}, \textit{tv}, \textit{lamp}, etc. Classes like \textit{wall}, \textit{floor}, and \textit{fridge} are the ones easily susceptible to misalignment on the tested dataset, albeit less than in $\mathcal{P}$.

\subsection{Robustness to Missing Geometric Information}

In this section, we provide an ablation study of \project{} on the 3D Scene Graph alignment task and evaluate how it performs on node alignment, when all the \textbf{geometric relationships} encoding positional information between the nodes such as \texttt{left} and \texttt{standing on} are missing. Results are in Table \ref{tab:missing_edges}. As expected, the structure module $\mathcal{S}$ suffers from this compared to the full ground-truth experiment, since the number of edges encoded in the neighborhood of an entity gets reduced. However, overall, our method does not show a drastic drop in node alignment metrics due to the absence of geometric relationships. This can be attributed to the fact that we do not discriminate between different types of relationships in our encoders, however, this is a very important robustness characteristic, especially, while working with predicted scene graphs where the relationships could be missing or incorrectly labelled. This also shows that once trained with full ground truth, our method is able to handle missing data during inference which would be useful for a navigation agent.

\begin{table}[ht!]
    \resizebox{\columnwidth}{!}{
    \centering
    \begin{tabular}{c|cccccc}
        \toprule
         \multirow{2}{*}{\textbf{Modalities}} & \textbf{Mean}  & \multicolumn{5}{c}{\textbf{Hits @ $\uparrow$}} \\
         & \textbf{RR $\uparrow$} & \textbf{K = 1} & \textbf{K = 2} & \textbf{K = 3} & \textbf{K = 4} & \textbf{K = 5} \\
        \midrule
        $\mathcal{P}$ & 0.884 & 0.835 & 0.886 & 0.921 & 0.938 & 0.951 \\
        $\mathcal{P}$ + $\mathcal{S}$ & 0.880 & 0.830 & 0.882 & 0.918 & 0.936 & 0.948 \\
        $\mathcal{P}$ + $\mathcal{S}$ + $\mathcal{R}$  & 0.893 & 0.844 & 0.898 & 0.933 & 0.949 & 0.959 \\
        \project{}  & \textbf{0.948} & \textbf{0.921} & \textbf{0.952} & \textbf{0.971} & \textbf{0.979} & \textbf{0.985} \\
        \bottomrule
    \end{tabular}
    }
    \vspace{1pt}
    \caption{\textbf{Evaluation on node matching.} We compare the performance of \project{} for different modality combinations, when all geometric edges are missing.}
    \label{tab:missing_edges}
\end{table}

\section{SGAligner Benchmark}
\label{sec:benchmark}
In this section, we offer qualitative explanations on our dataset generation procedure and discuss the evaluation metrics used to asses performance with respect to the various tasks we reported.

\begin{figure}[ht!]
    \centering
     \includegraphics[trim=0 0 0 0,clip,width=\columnwidth]{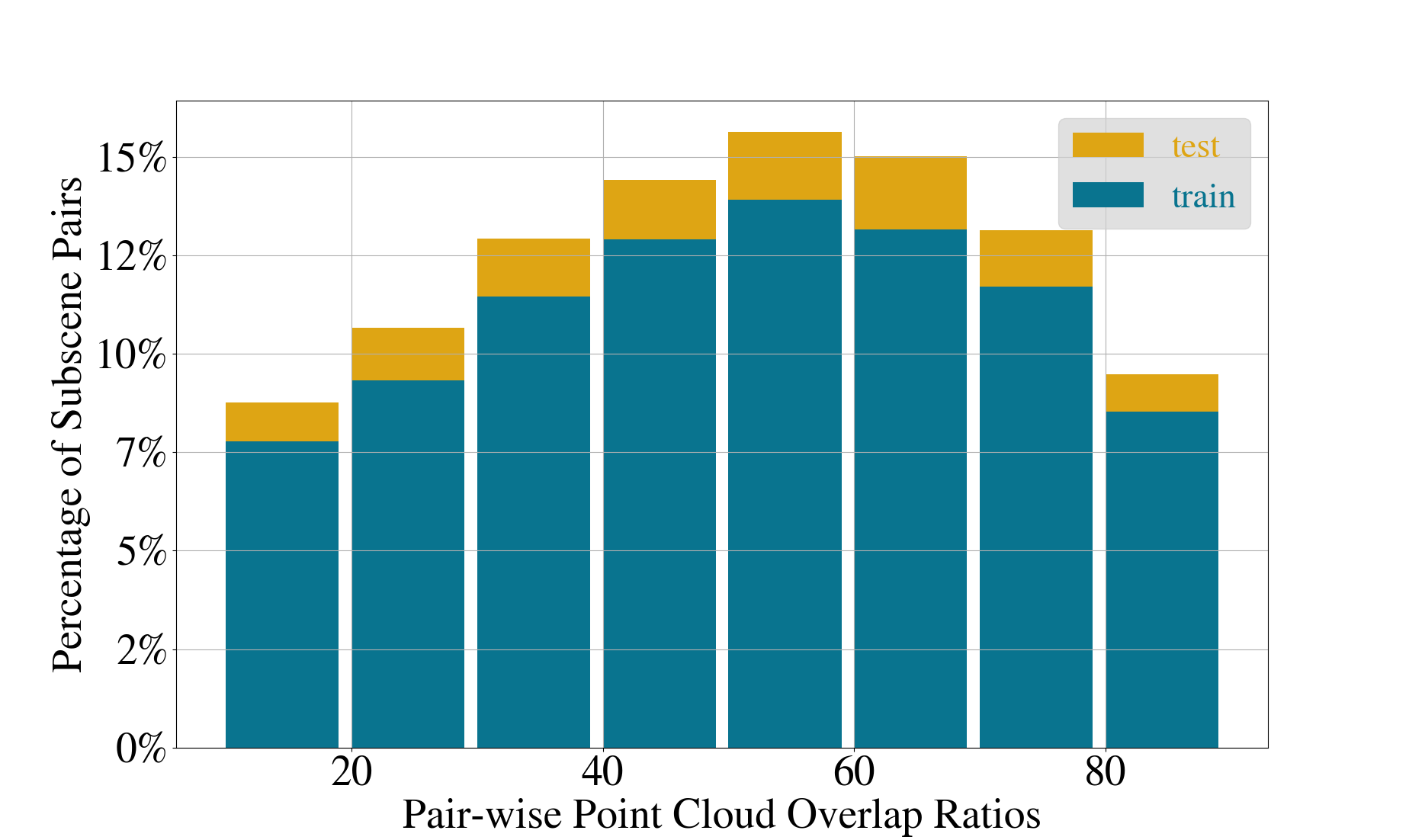}
    \caption{\textbf{Overlap statistics for the generated sub-scenes.}}
    \label{fig:overlap_subscene}
\end{figure} 

\subsection{Dataset}
\begin{figure*}[ht!]
    \centering
    \includegraphics[trim=0 0 0 5,clip,width=1.0\linewidth]{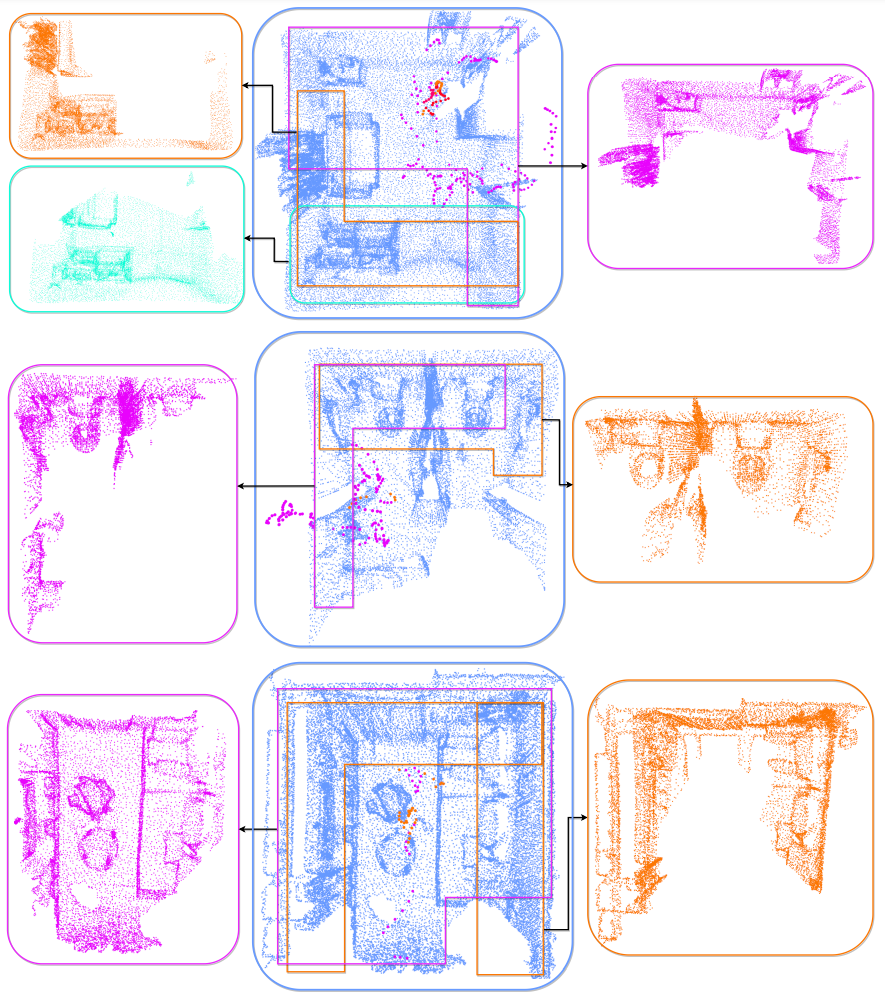}
    \caption{\textbf{Visualization of data generation process.} We visualise the creation of subscenes using our approach from Section 4 of the main paper. Given a point cloud from \cite{3rscan} (middle column), we create multiple sub-scenes (left and right columns) and showcase the used camera trajectory. The colors depict the camera trajectory and 3D spatial area of each sub-scene in the parent scene.}
    \label{fig:subscene_create}
\end{figure*}

In Sec. 4 of the main paper, we provide a description of the data generation procedure. In Figure \ref{fig:overlap_subscene}, we report statistics on the spatial overlap of the generated pairs. We show examples of sub-scenes generated using this approach in Figure \ref{fig:subscene_create}, alongside the camera trajectory used to capture the corresponding scan in \cite{3rscan}. We will make our code and benchmark public.

\subsection{Evaluation Metrics}
The evaluation metrics that we use to assess performance in Section 4.1 of the main paper, as well as in this
supplementary material, are formally defined in this section. 

\subsubsection{Alignment Metrics}
Inspired by works in multi-modal entity alignment \cite{mclea}, we define our alignment metrics as follows : \\

\textbf{Hits @ K} represents the fraction of true anchor entities present in the top k predictions :  
\begin{equation}
    H_k(r_1, ..., r_n) = \frac{1}{n} \sum_{i=1}^{n} I[r_i \le k] 
\end{equation}
where, $I[x \le y] = 1$ when $x \le y$ else $0$ and $k \in [1, 2, 3, 4, 5]$. \\

\textbf{Mean Reciprocal Rank (MRR)} corresponds to the arithmetic mean over the reciprocals of ranks of true triples. 

\begin{equation}
    MRR(r_1, ..., r_n) = \frac{1}{n}\sum_{i=1}^{n} r_i^{-1}
\end{equation}

\subsubsection{Registration Metrics}
\textbf{Feature Matching Recall (FMR)} \cite{predator}\cite{geotr} measures the fraction of point cloud pairs for which, based on the number of inlier correspondences, it is likely that accurate transformation parameters can be recovered with a robust estimator such as RANSAC.  It should be noted that FMR simply verifies whether the inlier ratio (IR) is higher than a threshold $\mathcal{T}$ = $0.05$. It does not examine if the transformation can actually be inferred from those correspondences, which is not always the case because of the possibility that their geometric arrangement is (almost) degenerate, such as when they are situated closely together or along a straight edge. 
\begin{equation}
    FMR = \frac{1}{M} \sum_{i=1}^{M}  \llbracket IR_i > \mathcal{T} \rrbracket
\end{equation}

\begin{equation}
    MRR(r_1, ..., r_n) = \frac{1}{n}\sum_{i=1}^{n} r_i^{-1}
\end{equation}
where $M$ is the number of all point cloud pairs.

\textbf{Registration Recall (RR)} is the fraction of registered point cloud pairs for which the transformation error is smaller than $0.2$m. The transformation error is the root mean squared error of the ground truth correspondence $\mathcal{H^*}$ after applying the predicted transformation $\textbf{T}_{P \rightarrow Q}$.

\begin{equation}
    RMSE = \sqrt{\frac{1}{|\mathcal{H^*}|} \sum_{({p^*}_{x_i}, {q^*}_{y_i}) \in \mathcal{H^*}} ||T_{P\rightarrow Q} 
           ({p^*}_{x_i}) - {q^*}_{y_i}||_2^2 }
\end{equation}

\begin{equation}
    RR = \frac{1}{M} \sum_{i=1}\llbracket RMSE_i < 0.2m \rrbracket
\end{equation}

\noindent
\textbf{Relative Rotation Error (RRE)} is the geodesic distance in degrees between estimated and ground-truth rotation matrices. 
\begin{equation}
    RRE = arccos ( \frac{trace({R^T} \cdot \Bar{R} - 1)}{2}  )
\end{equation}

\noindent
\textbf{Relative Translation Error (RTE)} is the the euclidean distance between estimated and ground-truth translation vectors. 
\begin{equation}
    RTE = || t - \Bar{t} ||    
\end{equation}

We compute mean RRE and RTE between all the registered point cloud pairs. 

\noindent
\textbf{Chamfer Distance} measures the quality of registration. Following \cite{rpmnet}, \cite{predator}, we use the 
\textit{modified} Chamfer distance metric : 

\begin{equation}
    \begin{aligned}
    \Bar{CD}(P, Q) = \frac{1}{|P|} \sum_{p \in P} min_{q \in Q_{raw}} || {T_P}^Q(p) - q||^2_2 + \\
                     \frac{1}{|Q|} \sum_{q \in Q} min_{p \in P_{raw}} || {q - T_P}^Q(p)||^2_2
    \end{aligned}
\end{equation}
where, $P_{raw}$ and $Q_{raw}$ are $raw/clean$ source and target point clouds respectively.

\subsubsection{Reconstruction Metrics}
The definition of full 3D reconstruction metrics is provided in Table \ref{tab:reconstruction_metrics}.
\begin{table}[ht!]
    \centering
    \begin{tabular}{c|c}
        \toprule
        \textbf{Metric} & \textbf{Definition} \\
        \midrule
        Acc & $mean_{p\in P}(min_{p^* \in P^*} || p - p^*||)$ \vspace{1pt} \\
        Comp & $mean_{p^* \in P^*}(min_{p \in P} || p - p^*||)$ \vspace{1pt} \\ 
        Precision & $mean_{p\in P}(min_{p^* \in P^*} || p - p^*|| < 0.05)$ \vspace{1pt} \\
        Recall & $mean_{p^* \in P^*}(min_{p \in P} || p - p^*|| < 0.05)$ \vspace{1pt} \\
        F1-Score & $\frac{2 * precision * recall}{precision + recall}$ \\
        \bottomrule
    \end{tabular}
    \vspace{3pt}
    \caption{\textbf{3D Reconstruction Metric Definitions.} $p$ and $p^*$ are ground truth and predicted point clouds respectively.}
    \label{tab:reconstruction_metrics}
\end{table}

\section{Implementation Details}
\label{sec:implementation}
Inspired by MCLEA \cite{mclea}, we use a multi-layered GAT with $2$ layers and each hidden unit being $128$-dimensional. All the modules output a $100$-dimensional embedding and the joint embedding, being a weighted concatenation, is $400$-dimensional. We use $\mathcal{T}_1$ for ICL loss as  $0.1$ and $\mathcal{T}_2$ for IAL loss as $1.0$. We train our model for $50$ epochs on a NVIDIA GeForce RTX 3060 Ti $8$GB GPU with a batch size of $4$ using AdamW \cite{loshchilov2018decoupled} optimizer and a learning rate of $0.001$.

{\small
\bibliographystyle{ieee_fullname}
\bibliography{main}

\begin{thebibliography}{10}\itemsep=-1pt

\bibitem{pmlr-v164-agia22a}
Christopher Agia, {Krishna Murthy} Jatavallabhula, Mohamed Khodeir, Ondrej Miksik, Vibhav Vineet, Mustafa Mukadam, Liam Paull, and Florian Shkurti.
\newblock Taskography: Evaluating robot task planning over large 3d scene graphs.
\newblock In {\em Conference on Robot Learning}, pages 46--58. PMLR, 2022.

\bibitem{armeni20193d}
Iro Armeni, Zhi-Yang He, JunYoung Gwak, Amir~R Zamir, Martin Fischer, Jitendra Malik, and Silvio Savarese.
\newblock 3d scene graph: A structure for unified semantics, 3d space, and camera.
\newblock In {\em Proceedings of the IEEE/CVF international conference on computer vision}, pages 5664--5673, 2019.

\bibitem{realtimepointnet}
Lin Bai, Yecheng Lyu, and Xinming Huang.
\newblock Pointnet on fpga for real-time lidar point cloud processing.
\newblock 10 2020.

\bibitem{d3feat}
Xuyang Bai, Zixin Luo, Lei Zhou, Hongbo Fu, Long Quan, and Chiew-Lan Tai.
\newblock D3feat: Joint learning of dense detection and description of 3d local features.
\newblock In {\em Proceedings of the IEEE/CVF conference on computer vision and pattern recognition}, pages 6359--6367, 2020.

\bibitem{barath2018graph}
Daniel Barath and Ji{\v{r}}{\'\i} Matas.
\newblock Graph-cut ransac.
\newblock In {\em Proceedings of the IEEE conference on computer vision and pattern recognition}, pages 6733--6741, 2018.

\bibitem{barath2020magsac++}
Daniel Barath, Jana Noskova, Maksym Ivashechkin, and Jiri Matas.
\newblock Magsac++, a fast, reliable and accurate robust estimator.
\newblock In {\em Proceedings of the IEEE/CVF conference on computer vision and pattern recognition}, pages 1304--1312, 2020.

\bibitem{dlite}
Yun Chang, Luca Ballotta, and Luca Carlone.
\newblock D-lite: Navigation-oriented compression of 3d scene graphs under communication constraints.
\newblock {\em arXiv preprint arXiv:2209.06111}, 2022.

\bibitem{mmea}
Liyi Chen, Zhi Li, Yijun Wang, Tong Xu, Zhefeng Wang, and Enhong Chen.
\newblock Mmea: entity alignment for multi-modal knowledge graph.
\newblock In {\em Knowledge Science, Engineering and Management: 13th International Conference, KSEM 2020, Hangzhou, China, August 28--30, 2020, Proceedings, Part I 13}, pages 134--147. Springer, 2020.

\bibitem{chen2022multi}
Liyi Chen, Zhi Li, Tong Xu, Han Wu, Zhefeng Wang, Nicholas~Jing Yuan, and Enhong Chen.
\newblock Multi-modal siamese network for entity alignment.
\newblock In {\em Proceedings of the 28th ACM SIGKDD Conference on Knowledge Discovery and Data Mining}, pages 118--126, 2022.

\bibitem{cheng2022multijaf}
Bo Cheng, Jia Zhu, and Meimei Guo.
\newblock Multijaf: Multi-modal joint entity alignment framework for multi-modal knowledge graph.
\newblock {\em Neurocomputing}, 500:581--591, 2022.

\bibitem{dhamo2021graph}
Helisa Dhamo, Fabian Manhardt, Nassir Navab, and Federico Tombari.
\newblock Graph-to-3d: End-to-end generation and manipulation of 3d scenes using scene graphs.
\newblock In {\em Proceedings of the IEEE/CVF International Conference on Computer Vision}, pages 16352--16361, 2021.

\bibitem{martin1981random}
Martin~A. Fischler and Robert~C. Bolles.
\newblock Random sample consensus: A paradigm for model fitting with applications to image analysis and automated cartography.
\newblock {\em Communication of ACM}, 1981.

\bibitem{Gadre_2022_CVPR}
Samir~Yitzhak Gadre, Kiana Ehsani, Shuran Song, and Roozbeh Mottaghi.
\newblock Continuous scene representations for embodied ai.
\newblock In {\em Proceedings of the IEEE/CVF Conference on Computer Vision and Pattern Recognition (CVPR)}, pages 14849--14859, June 2022.

\bibitem{hmea}
Hao Guo, Jiuyang Tang, Weixin Zeng, Xiang Zhao, and Li Liu.
\newblock Multi-modal entity alignment in hyperbolic space.
\newblock {\em Neurocomputing}, 461:598--607, 2021.

\bibitem{pct2021}
Meng-Hao Guo, Jun-Xiong Cai, Zheng-Ning Liu, Tai-Jiang Mu, Ralph~R. Martin, and Shi-Min Hu.
\newblock Pct: Point cloud transformer.
\newblock {\em Computational Visual Media}, 7(2):187–199, Apr 2021.

\bibitem{predator}
Shengyu Huang, Zan Gojcic, Mikhail Usvyatsov, Andreas Wieser, and Konrad Schindler.
\newblock Predator: Registration of 3d point clouds with low overlap.
\newblock In {\em Proceedings of the IEEE/CVF conference on computer vision and pattern recognition}, 2021.

\bibitem{hydra}
N. Hughes, Y. Chang, and L. Carlone.
\newblock Hydra: A real-time spatial perception system for {3D} scene graph construction and optimization.
\newblock 2022.

\bibitem{vlad}
Herv{\'e} J{\'e}gou, Matthijs Douze, Cordelia Schmid, and Patrick P{\'e}rez.
\newblock Aggregating local descriptors into a compact image representation.
\newblock In {\em 2010 IEEE computer society conference on computer vision and pattern recognition}, pages 3304--3311. IEEE, 2010.

\bibitem{jiao2022sequential}
Ziyuan Jiao, Yida Niu, Zeyu Zhang, Song-Chun Zhu, Yixin Zhu, and Hangxin Liu.
\newblock Sequential manipulation planning on scene graph.
\newblock In {\em 2022 IEEE/RSJ International Conference on Intelligent Robots and Systems (IROS)}, pages 8203--8210. IEEE, 2022.

\bibitem{kim20193}
Ue-Hwan Kim, Jin-Man Park, Taek-Jin Song, and Jong-Hwan Kim.
\newblock 3-d scene graph: A sparse and semantic representation of physical environments for intelligent agents.
\newblock {\em IEEE transactions on cybernetics}, 50(12):4921--4933, 2019.

\bibitem{krishna2017visual}
Ranjay Krishna, Yuke Zhu, Oliver Groth, Justin Johnson, Kenji Hata, Joshua Kravitz, Stephanie Chen, Yannis Kalantidis, Li-Jia Li, David~A Shamma, et~al.
\newblock Visual genome: Connecting language and vision using crowdsourced dense image annotations.
\newblock {\em International journal of computer vision}, 123:32--73, 2017.

\bibitem{li2022embodied}
Xinghang Li, Di Guo, Huaping Liu, and Fuchun Sun.
\newblock Embodied semantic scene graph generation.
\newblock In {\em Conference on Robot Learning}, pages 1585--1594. PMLR, 2022.

\bibitem{li2019graph}
Yujia Li, Chenjie Gu, Thomas Dullien, Oriol Vinyals, and Pushmeet Kohli.
\newblock Graph matching networks for learning the similarity of graph structured objects.
\newblock In {\em International conference on machine learning}, pages 3835--3845. PMLR, 2019.

\bibitem{li2022remote}
Yongwei Li, Yalong Ma, Xiang Huo, and Xinkai Wu.
\newblock Remote object navigation for service robots using hierarchical knowledge graph in human-centered environments.
\newblock {\em Intelligent Service Robotics}, 15(4):459--473, 2022.

\bibitem{mclea}
Zhenxi Lin, Ziheng Zhang, Meng Wang, Yinghui Shi, Xian Wu, and Yefeng Zheng.
\newblock Multi-modal contrastive representation learning for entity alignment.
\newblock {\em arXiv preprint arXiv:2209.00891}, 2022.

\bibitem{eva}
Fangyu Liu, Muhao Chen, Dan Roth, and Nigel Collier.
\newblock Visual pivoting for (unsupervised) entity alignment.
\newblock In {\em Proceedings of the AAAI Conference on Artificial Intelligence}, volume~35, pages 4257--4266, 2021.

\bibitem{hal}
Fangyu Liu, Rongtian Ye, Xun Wang, and Shuaipeng Li.
\newblock Hal: Improved text-image matching by mitigating visual semantic hubs.
\newblock {\em Proceedings of the AAAI Conference on Artificial Intelligence}, 34:11563--11571, 04 2020.

\bibitem{mmkg}
Ye Liu, Hui Li, Alberto Garcia-Duran, Mathias Niepert, Daniel Onoro-Rubio, and David~S Rosenblum.
\newblock Mmkg: multi-modal knowledge graphs.
\newblock In {\em The Semantic Web: 16th International Conference, ESWC 2019, Portoro{\v{z}}, Slovenia, June 2--6, 2019, Proceedings 16}, pages 459--474. Springer, 2019.

\bibitem{3dvsg}
Samuel Looper, Javier Rodriguez-Puigvert, Roland Siegwart, Cesar Cadena, and Lukas Schmid.
\newblock 3d vsg: Long-term semantic scene change prediction through 3d variable scene graphs.
\newblock {\em arXiv preprint arXiv:2209.07896}, 2022.

\bibitem{loshchilov2018decoupled}
Ilya Loshchilov and Frank Hutter.
\newblock Decoupled weight decay regularization.
\newblock In {\em International Conference on Learning Representations}, 2019.

\bibitem{qi2017pointnet}
Charles~R Qi, Hao Su, Kaichun Mo, and Leonidas~J Guibas.
\newblock Pointnet: Deep learning on point sets for 3d classification and segmentation.
\newblock In {\em Proceedings of the IEEE conference on computer vision and pattern recognition}, pages 652--660, 2017.

\bibitem{geotr}
Zheng Qin, Hao Yu, Changjian Wang, Yulan Guo, Yuxing Peng, and Kai Xu.
\newblock Geometric transformer for fast and robust point cloud registration.
\newblock In {\em Proceedings of the IEEE/CVF conference on computer vision and pattern recognition}, 2022.

\bibitem{raguram2012usac}
Rahul Raguram, Ondrej Chum, Marc Pollefeys, Jiri Matas, and Jan-Michael Frahm.
\newblock Usac: A universal framework for random sample consensus.
\newblock {\em IEEE transactions on pattern analysis and machine intelligence}, 35(8):2022--2038, 2012.

\bibitem{ravichandran2020bridging}
Zachary Ravichandran, J~Daniel Griffith, Benjamin Smith, and Costas Frost.
\newblock Bridging scene understanding and task execution with flexible simulation environments.
\newblock {\em arXiv preprint arXiv:2011.10452}, 2020.

\bibitem{9812179}
Zachary Ravichandran, Lisa Peng, Nathan Hughes, J.~Daniel Griffith, and Luca Carlone.
\newblock Hierarchical representations and explicit memory: Learning effective navigation policies on 3d scene graphs using graph neural networks.
\newblock In {\em 2022 International Conference on Robotics and Automation (ICRA)}, pages 9272--9279, 2022.

\bibitem{rosinol20203d}
Antoni Rosinol, Arjun Gupta, Marcus Abate, Jingnan Shi, and Luca Carlone.
\newblock 3d dynamic scene graphs: Actionable spatial perception with places, objects, and humans.
\newblock {\em arXiv preprint arXiv:2002.06289}, 2020.

\bibitem{rosinol2021kimera}
Antoni Rosinol, Andrew Violette, Marcus Abate, Nathan Hughes, Yun Chang, Jingnan Shi, Arjun Gupta, and Luca Carlone.
\newblock Kimera: From slam to spatial perception with 3d dynamic scene graphs.
\newblock {\em The International Journal of Robotics Research}, 40(12-14):1510--1546, 2021.

\bibitem{fpfh}
Radu~Bogdan Rusu, Nico Blodow, and Michael Beetz.
\newblock Fast point feature histograms (fpfh) for 3d registration.
\newblock In {\em 2009 IEEE international conference on robotics and automation}, pages 3212--3217. IEEE, 2009.

\bibitem{sepulveda2018deep}
Gabriel Sepulveda, Juan~Carlos Niebles, and Alvaro Soto.
\newblock A deep learning based behavioral approach to indoor autonomous navigation.
\newblock In {\em 2018 IEEE international conference on robotics and automation (ICRA)}, pages 4646--4653. IEEE, 2018.

\bibitem{masked-mmea}
Yinghui Shi, Meng Wang, Ziheng Zhang, Zhenxi Lin, and Yefeng Zheng.
\newblock Probing the impacts of visual context in multimodal entity alignment.
\newblock In {\em Web and Big Data: 6th International Joint Conference, APWeb-WAIM 2022, Nanjing, China, November 25--27, 2022, Proceedings, Part II}, pages 255--270. Springer, 2023.

\bibitem{sun2021neucon}
Jiaming Sun, Yiming Xie, Linghao Chen, Xiaowei Zhou, and Hujun Bao.
\newblock {NeuralRecon}: Real-time coherent {3D} reconstruction from monocular video.
\newblock {\em CVPR}, 2021.

\bibitem{velikovi2018graph}
Petar Veličković, Guillem Cucurull, Arantxa Casanova, Adriana Romero, Pietro Liò, and Yoshua Bengio.
\newblock Graph attention networks.
\newblock In {\em International Conference on Learning Representations}, 2018.

\bibitem{3rscan}
Johanna Wald, Armen Avetisyan, Nassir Navab, Federico Tombari, and Matthias Nie{\ss}ner.
\newblock Rio: 3d object instance re-localization in changing indoor environments.
\newblock In {\em Proceedings of the IEEE/CVF International Conference on Computer Vision}, pages 7658--7667, 2019.

\bibitem{3dssg}
Johanna Wald, Helisa Dhamo, Nassir Navab, and Federico Tombari.
\newblock Learning 3d semantic scene graphs from 3d indoor reconstructions.
\newblock In {\em Proceedings of the IEEE/CVF Conference on Computer Vision and Pattern Recognition}, pages 3961--3970, 2020.

\bibitem{wu2021scenegraphfusion}
Shun-Cheng Wu, Johanna Wald, Keisuke Tateno, Nassir Navab, and Federico Tombari.
\newblock Scenegraphfusion: Incremental 3d scene graph prediction from rgb-d sequences.
\newblock In {\em Proceedings of the IEEE/CVF Conference on Computer Vision and Pattern Recognition}, pages 7515--7525, 2021.

\bibitem{rpmnet}
Zi~Jian Yew and Gim~Hee Lee.
\newblock Rpm-net: Robust point matching using learned features.
\newblock In {\em Proceedings of the IEEE/CVF conference on computer vision and pattern recognition}, pages 11824--11833, 2020.

\bibitem{regtr}
Zi~Jian Yew and Gim~Hee Lee.
\newblock Regtr: End-to-end point cloud correspondences with transformers.
\newblock In {\em Proceedings of the IEEE/CVF conference on computer vision and pattern recognition}, 2022.

\bibitem{3dmatch}
Andy Zeng, Shuran Song, Matthias Nie{\ss}ner, Matthew Fisher, Jianxiong Xiao, and Thomas Funkhouser.
\newblock 3dmatch: Learning local geometric descriptors from rgb-d reconstructions.
\newblock In {\em CVPR}, 2017.

\bibitem{SGGpoint}
Chaoyi Zhang, Jianhui Yu, Yang Song, and Weidong Cai.
\newblock Exploiting edge-oriented reasoning for 3d point-based scene graph analysis.
\newblock In {\em IEEE/CVF Conference on Computer Vision and Pattern Recognition (CVPR)}, pages 9705--9715, June 2021.

\bibitem{mkhan}
Qingheng Zhang, Zequn Sun, Wei Hu, Muhao Chen, Lingbing Guo, and Yuzhong Qu.
\newblock Multi-view knowledge graph embedding for entity alignment.
\newblock {\em arXiv preprint arXiv:1906.02390}, 2019.

\bibitem{zhangdual}
Ruohan Zhang, Dhruva Bansal, Yilun Hao, Ayano Hiranaka, Jialu Gao, Chen Wang, Roberto Mart{\'\i}n-Mart{\'\i}n, Li Fei-Fei, and Jiajun Wu.
\newblock A dual representation framework for robot learning with human guidance.
\newblock In {\em 6th Annual Conference on Robot Learning}.

\bibitem{knowledgescene}
Shoulong Zhang, Shuai Li, Aimin Hao, and Hong Qin.
\newblock Knowledge-inspired 3d scene graph prediction in point cloud.
\newblock In Marc'Aurelio Ranzato, Alina Beygelzimer, Yann~N. Dauphin, Percy Liang, and Jennifer~Wortman Vaughan, editors, {\em Advances in Neural Information Processing Systems 34: Annual Conference on Neural Information Processing Systems 2021, NeurIPS 2021, December 6-14, 2021, virtual}, pages 18620--18632, 2021.

\bibitem{iss}
Yu Zhong.
\newblock Intrinsic shape signatures: A shape descriptor for 3d object recognition.
\newblock In {\em 2009 IEEE 12th international conference on computer vision workshops, ICCV workshops}, pages 689--696. IEEE, 2009.

\end{thebibliography}
}

\end{document}